\documentclass{article}
\usepackage[preprint]{neurips_2020}
\newcommand{\xs}{\scriptsize}
\usepackage{geometry}
\usepackage{graphicx}
\usepackage{qtree}
\usepackage{tikz}
\usetikzlibrary{positioning,backgrounds,fit}
\usepackage{tabularx}
\usepackage{booktabs}
\usepackage{siunitx}
\usepackage{ragged2e}
\usepackage{rotating}
\usepackage{adjustbox}
\usepackage{caption}
\usepackage[T1]{fontenc}
\usepackage{pdflscape}
\usepackage{nameref}
\usepackage{pdfpages}
\usepackage[multiple]{footmisc}
\usepackage{float}
\newcolumntype{R}[1]{>{\RaggedRight}p{#1}}
\usepackage{amssymb}
\usepackage[x11names,dvipsnames]{xcolor}
\usepackage{verbatim}
\usepackage{subcaption}
\usepackage{array}
\usepackage{natbib}
\usepackage{hyperref}
\usepackage{mathtools}
\usepackage{xspace}
\usepackage{cleveref}
\usepackage{svg}
\usepackage{transparent}
\usepackage[toc,title,page]{appendix}
\definecolor{orangefull}{RGB}{230, 159, 0}
\newcommand*{\eg}{e.g.\@\xspace}

\usepackage{siunitx}
\usetikzlibrary{shapes.geometric, positioning, arrows.meta, fit}

\usepackage[pagewise]{lineno}  
\title{A hybrid machine learning-analog framework for downscaling unprecedented climate extremes}
\author{%
  Nina Effenberger$^{1}$\thanks{Corresponding author: \textit{nina.effenberger@env.ethz.ch}} \quad 
  Florian E. Roemer$^{1}$ \quad 
  Maybritt Schillinger$^{2}$\\
  \textbf{Maxim Samarin$^{3}$} \quad 
  \textbf{Erich Fischer}$^{1}$ \quad 
  \textbf{Reto Knutti}$^{1}$ \\[1ex]
  $^{1}$Institute for Atmospheric and Climate Science, ETH Zurich \\
  $^{2}$Seminar for Statistics, ETH Zurich \quad $^{3}$Swiss Data Science Center, EPFL and ETH Zurich
}

\begin{document}

\maketitle

\begin{abstract}
Extreme weather events are critical for climate impact assessments. While physical models can simulate unprecedented extremes, for example, through ensemble boosting, this approach is computationally demanding, often restricting output to coarse spatial resolutions. Probabilistic generative downscaling offers a solution by enabling efficient sampling from the entire climate distribution, including its extreme tails. Here, we first demonstrate that machine learning generated regional climate model heatwaves over Paris are physically realistic. We then leverage an emulated machine learning dataset to perform analog-based downscaling of coarse global climate model ensemble boosting events. Our hybrid framework effectively bridges high-resolution regional detail with rare-event sampling, providing an efficient tool for extreme event downscaling and capturing multi-model uncertainty.
\end{abstract}

\section{Introduction}
Despite the computational effort dedicated to multi-model initiatives like the Coupled Model Intercomparison Project CMIP \citep{eyring2016overview}, their generally coarse spatial scale limits their utility for impact assessments at local to regional scale, in urbanized areas, over complex topography, and along coastlines. Evaluating more localized risks related to heatwaves and compound events benefits from fine-scale detail \citep{clarke2022extreme}, yet practical analyses remain limited by the inherent scarcity of extreme events and the spatial resolution of standard global climate model outputs.
To overcome the sample-size constraints, rare event algorithms such as ensemble boosting \citep{gessner2021very} and the GKTL algorithm \citep{giardina2011simulating} have emerged as tools for generating physically coherent extremes. Ensemble boosting addresses the sample-size constraints by perturbing the initial atmospheric conditions of targeted  events days or weeks in advance. It efficiently samples low-probability, high-impact extremes while maintaining physical coherence within climate models. Yet, despite its conceptual efficiency, generating these large ensembles is computationally intensive, particularly in higher-resolution climate models \citep{balaji2017cpmip}.

Dynamical downscaling of these boosted global climate samples presents an even greater technical and computational challenge. Running dynamical regional climate models over hundreds of boosted ensemble members to obtain high-resolution fields is computationally prohibitive. Consequently, extreme event storyline studies frequently rely directly on global climate model outputs \citep{fischer_storylines_2023}. While statistical downscaling is possible, statistical approaches struggle with unprecedented extremes, where observational analogs based on historical archives do not exist for future record-shattering extremes that exceed the historic range \citep{noyelle2025attributing}.

To bridge the gap between low-resolution ensemble-boosted simulations and fine-scale detail, we leverage machine learning emulators to sample climate extremes. Our approach combines physics-based GCM ensemble boosting with a general-purpose, temporally-consistent machine learning (ML) downscaling emulator to construct a large, physically consistent high-resolution climate dataset. By combining high-resolution ML-generated data with low-resolution physical extremes, coarse GCM extremes can be mapped to high-resolution regional states, and climate model uncertainty can be used to construct alternative storylines of extreme events. 

We demonstrate the utility of this framework using Western European heatwaves studied in \citet{fischer_storylines_2023}, centered on Paris, as a primary proof-of-concept. More specifically, by training the temporally consistent ML emulator \textit{EnScale} \citep{schillinger2026enscale} on physics-based GCM-RCM simulations, we create a library of physically consistent, fine-scale climate emulations. The emulator's computational efficiency allows us to sample 8,000 summers, including extremes, across 8 different GCM-RCM combinations. We show that these ML-generated samples span the range of the most extreme boosted events, allowing us to perform analog matching to instantaneously downscale coarse boosted GCM events. Ultimately, our hybrid framework preserves spatio-temporal consistency while capturing multi-model variability at a fraction of the computational cost of traditional dynamical downscaling.

\section{Methods}
\subsection{Framework overview}
\begin{figure}[H]
    \centering
\begin{tikzpicture}[
    node distance=0.5cm, 
    block/.style={rectangle, draw=gray, dashed, align=center, inner sep=1pt}, 
    noborder/.style={rectangle, align=center, inner sep=1pt}                   
]

    \def\singleImgWidth{2.7cm}
    \def\tripletImgWidth{1.4cm}

    \setlength{\tabcolsep}{1pt}

    \node[noborder] (gcm) {
        \includegraphics[width=\singleImgWidth]{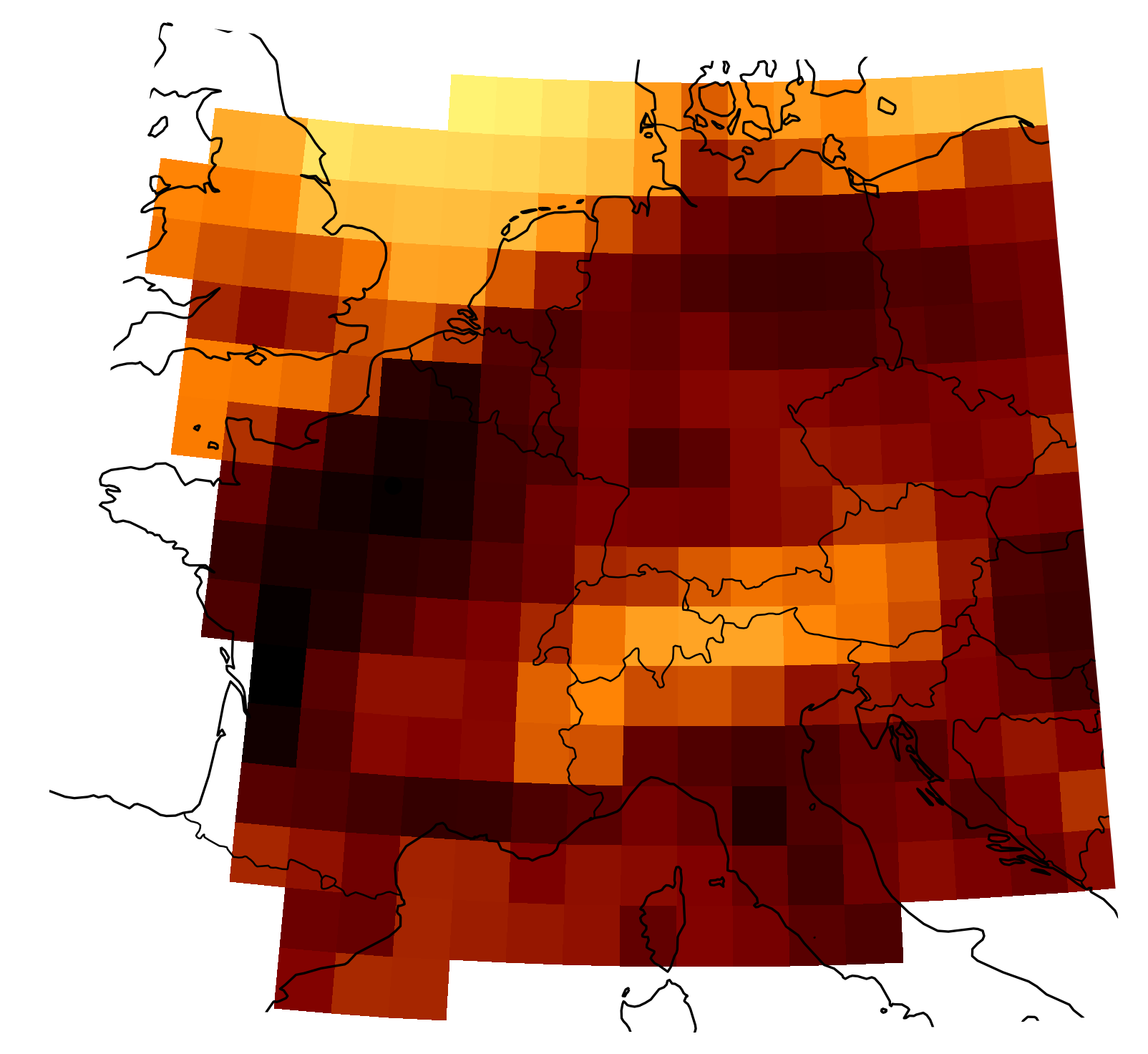}
    };

    \node[block] (boost) [below=of gcm] {
        \begin{tabular}{ccc}
            \includegraphics[width=\tripletImgWidth]{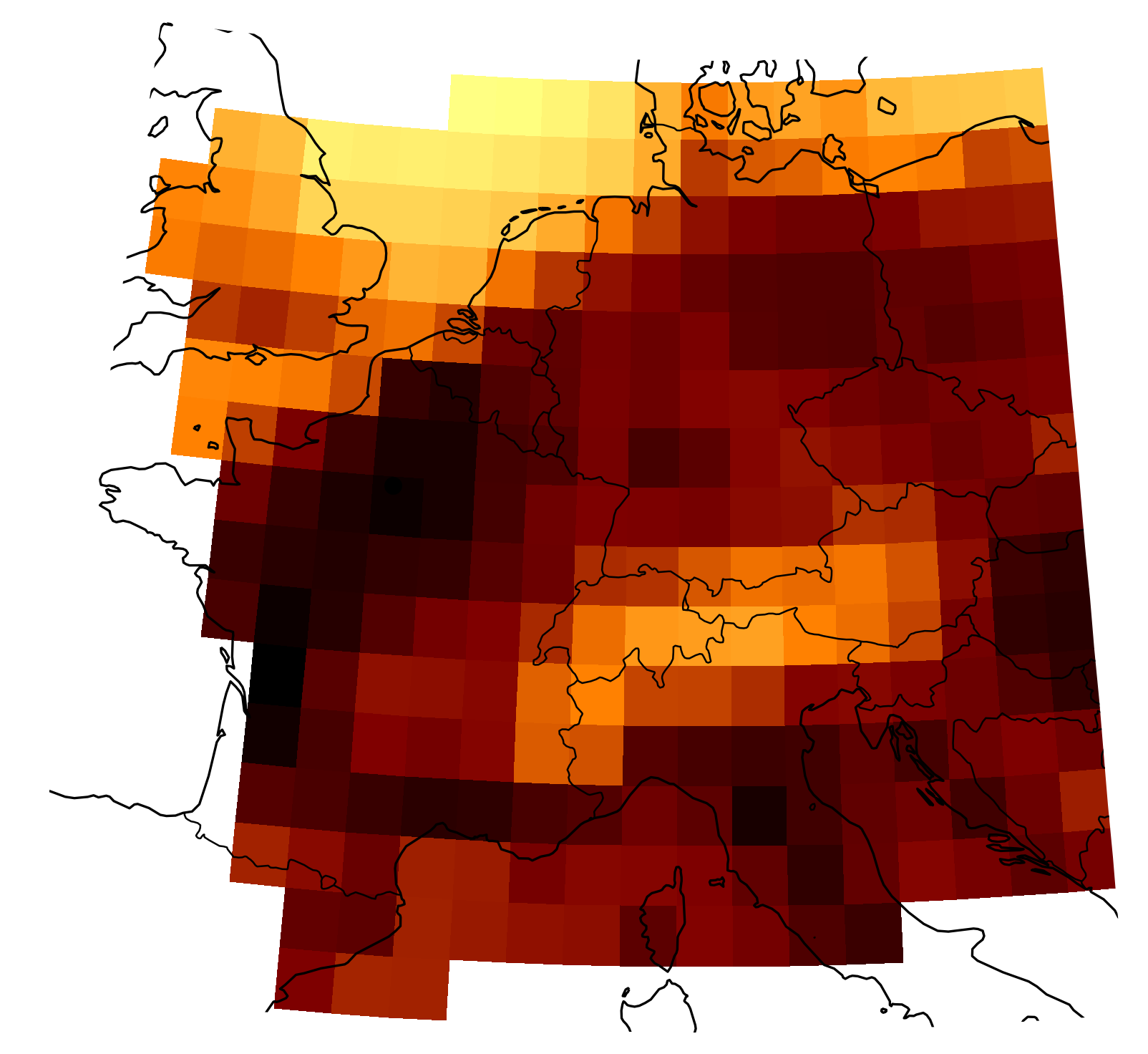} & 
            \includegraphics[width=\tripletImgWidth]{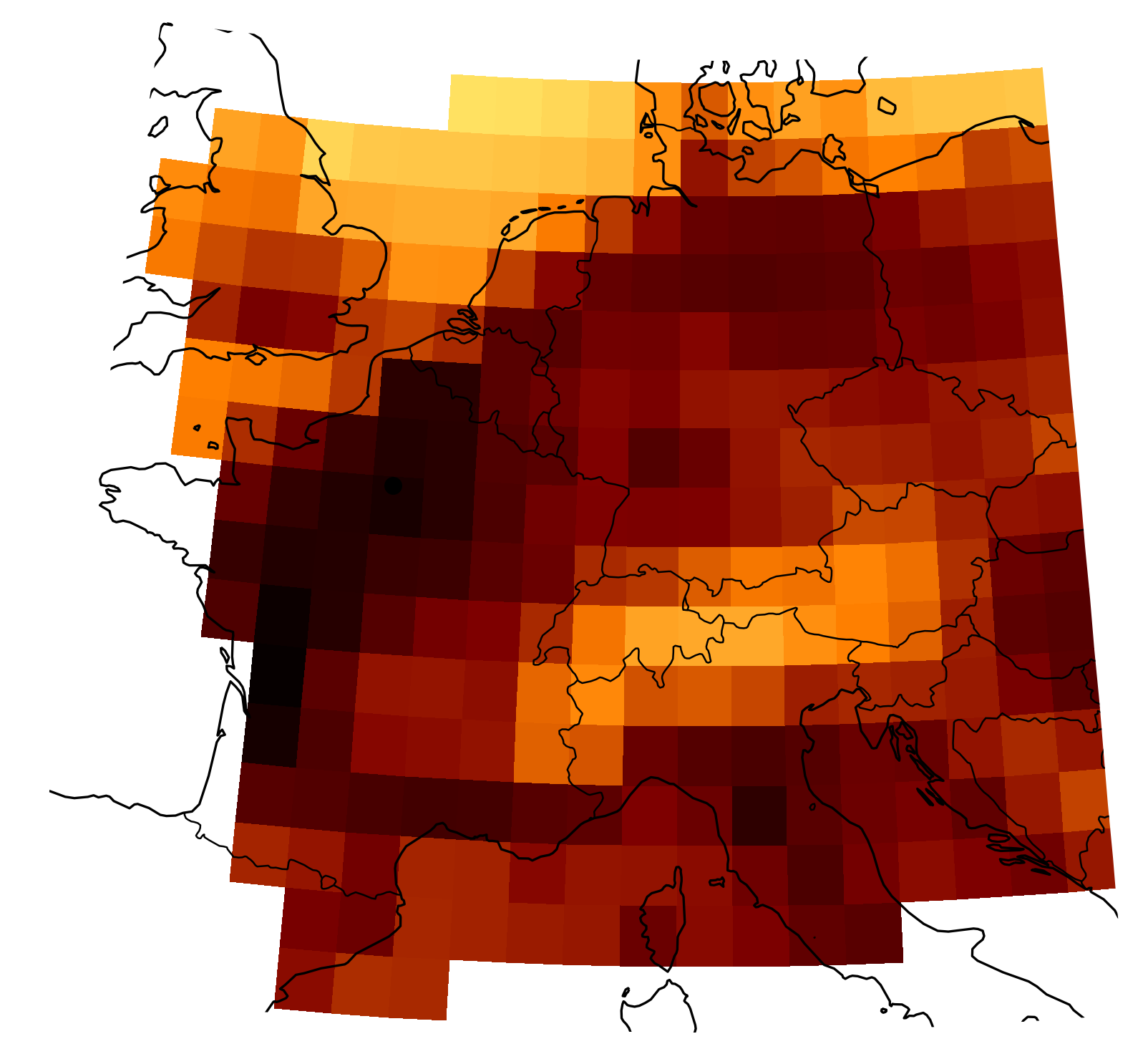} & 
            \includegraphics[width=\tripletImgWidth]{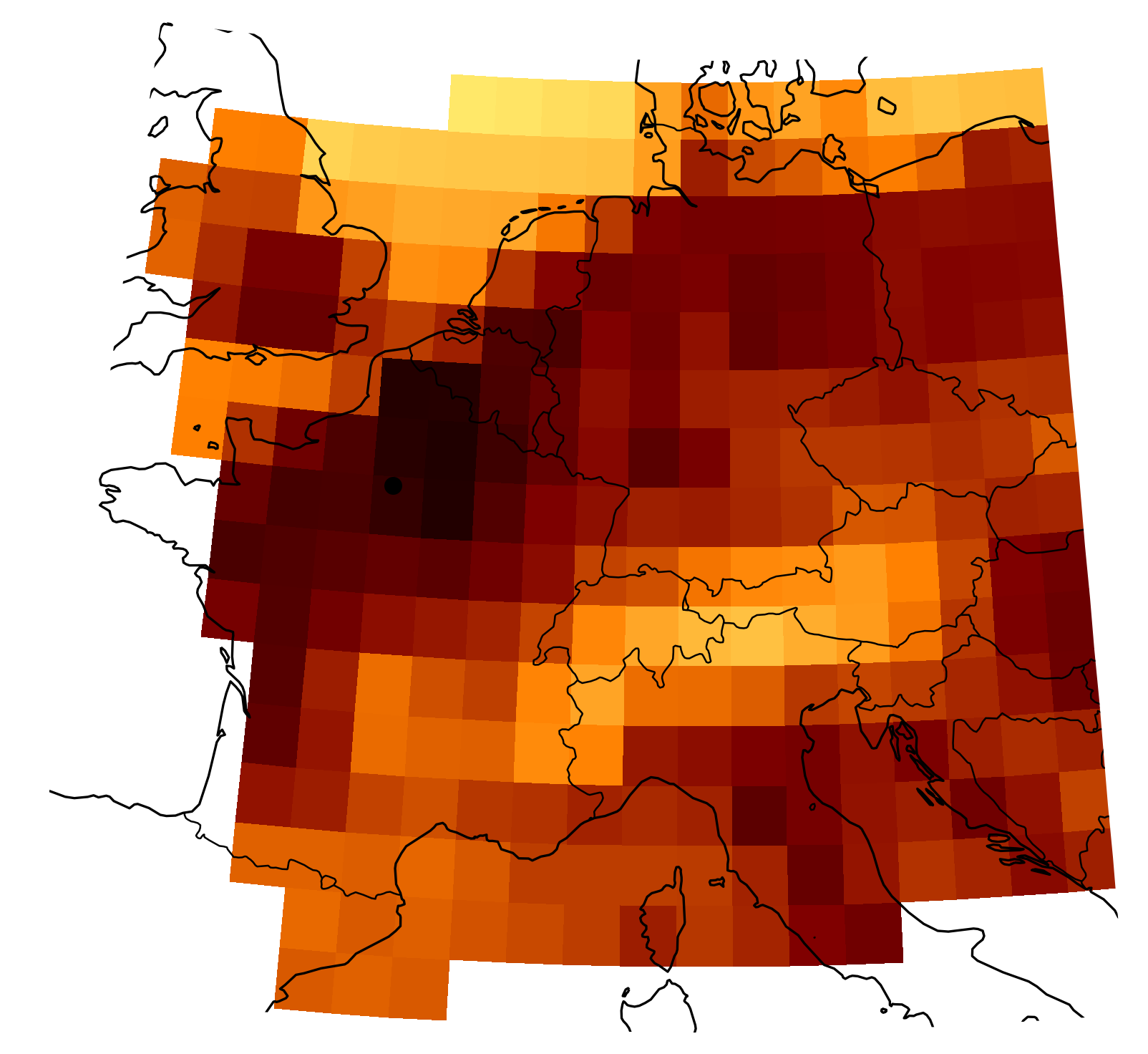} 
        \end{tabular}
    };

    \node[noborder] (rcm) [right=3.2cm of gcm] {
        \includegraphics[width=\singleImgWidth]{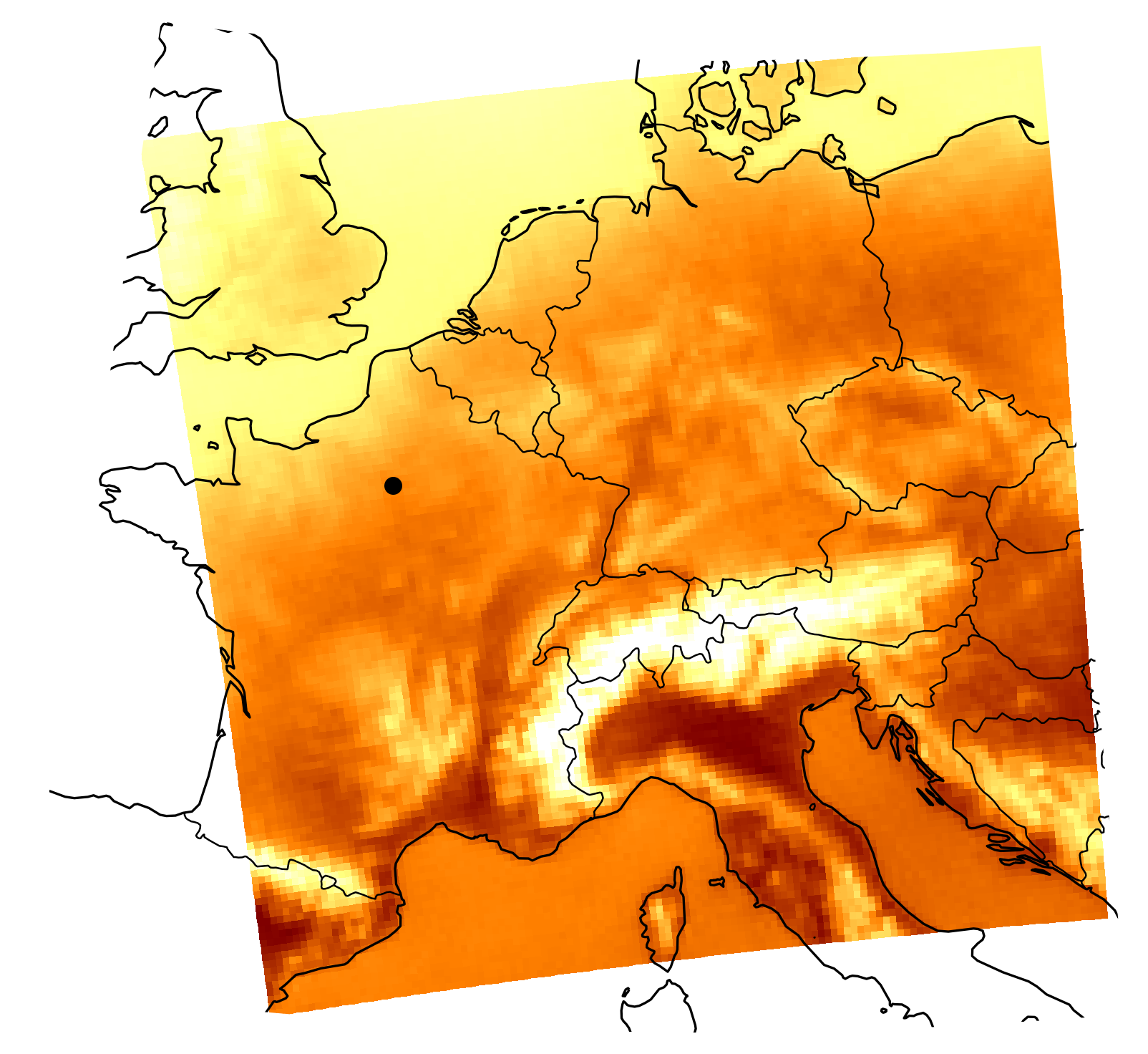}
    };

    \node[block] (emul) [below=of rcm] {
        \begin{tabular}{ccc}
            \includegraphics[width=\tripletImgWidth]{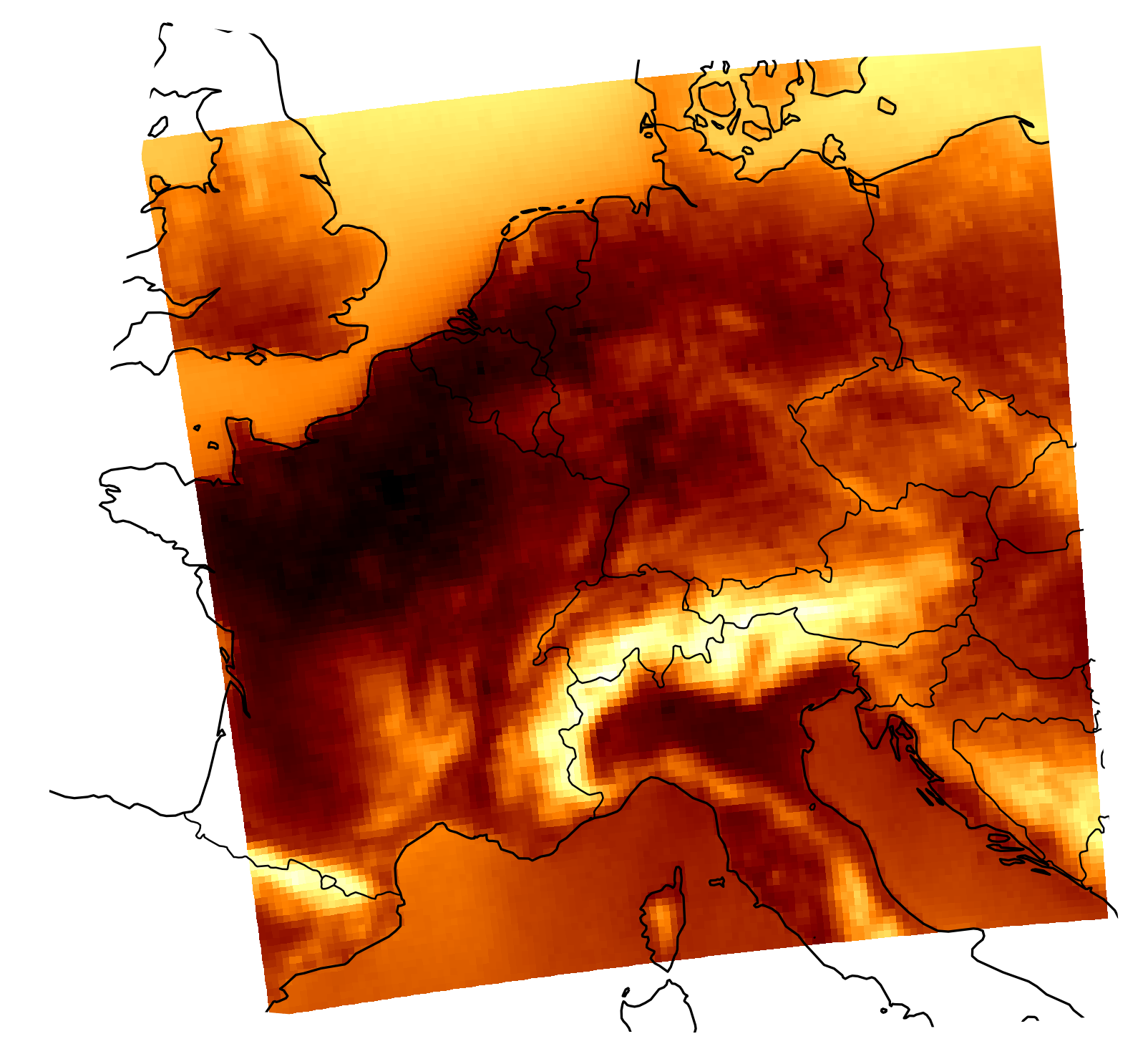} & 
            \includegraphics[width=\tripletImgWidth]{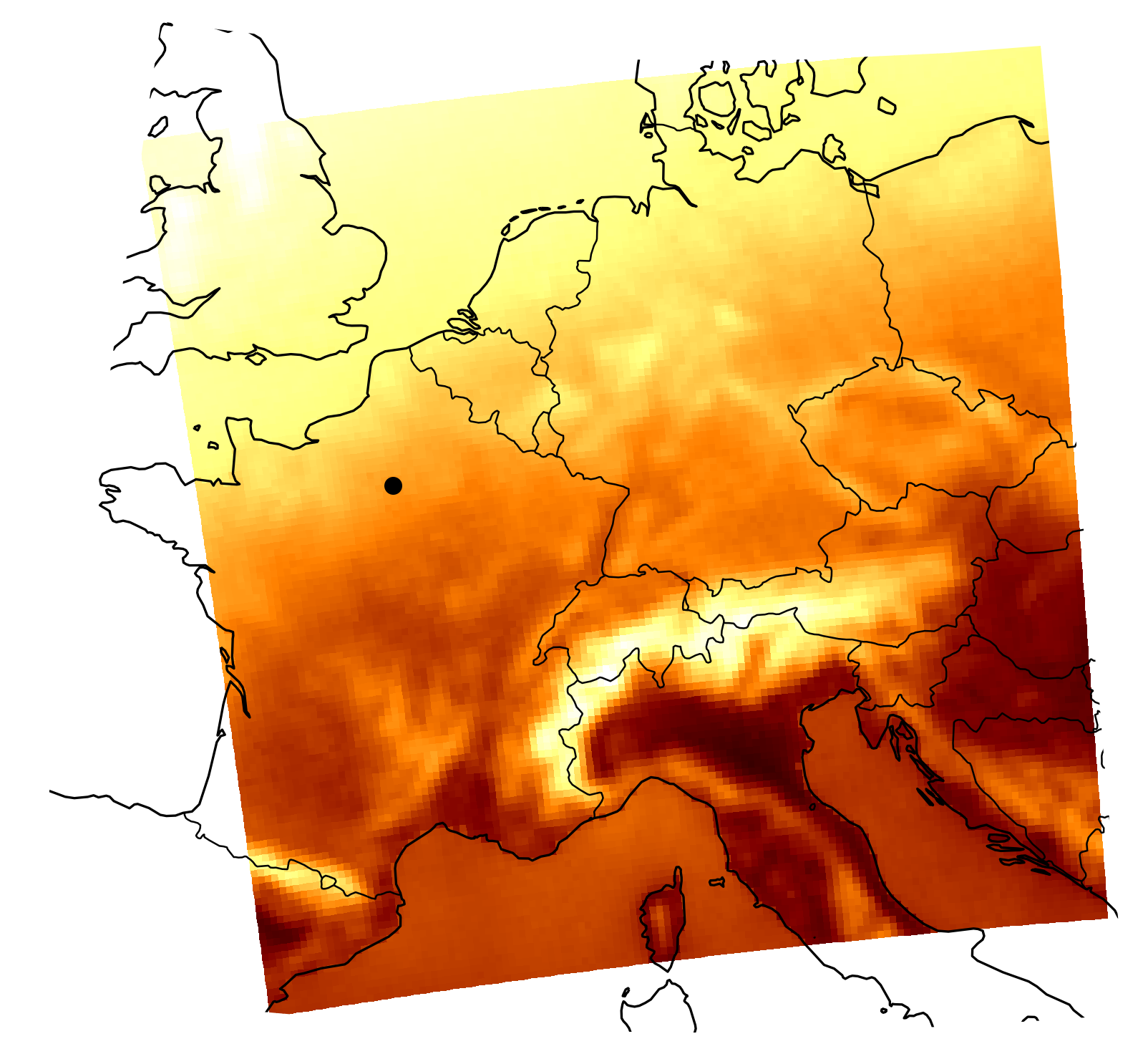} & 
            \includegraphics[width=\tripletImgWidth]{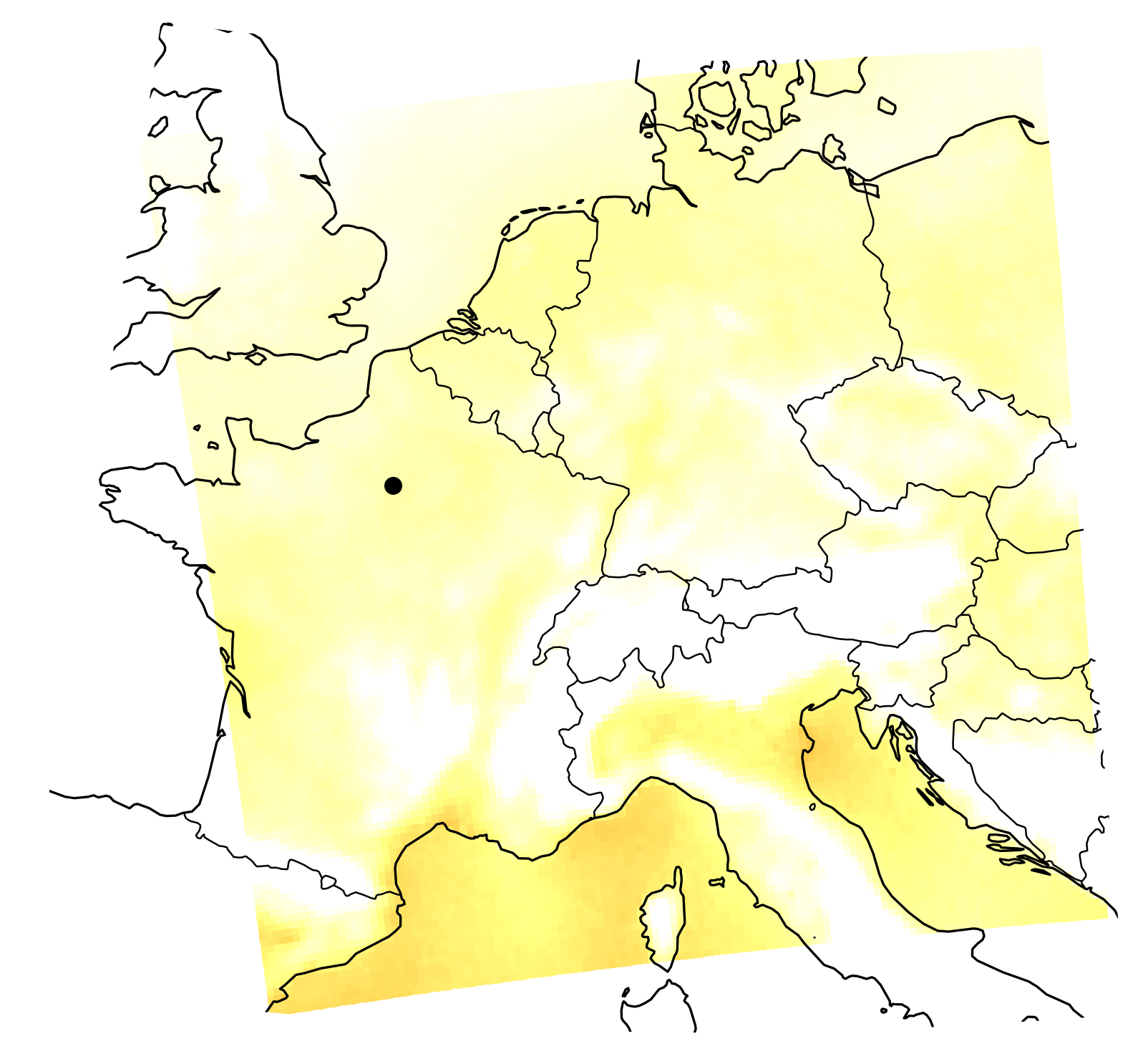}
        \end{tabular}
    };

    \node[above=0.2cm of gcm, font=\footnotesize\bfseries] {Ensemble boosting};
    \node[above=0.2cm of rcm, font=\footnotesize\bfseries] {ML emulation};

    \draw[->, thick] (gcm) -- (boost) node[midway, right, font=\xs] {boost};

    \draw[->, thick] (rcm) -- (emul) node[midway, right, font=\xs] {emulate};

    \draw[->, thick] (boost) -- (emul) node[midway, below, align=center, font=\xs] {analog \\ matching};
\end{tikzpicture}
    \caption{Overview of the hybrid extreme-event framework. Low-resolution ensemble boosting leverages a single extreme event as an initial condition to generate a large sample of physically consistent extremes with a global climate model (darker shades indicate higher temperatures). In parallel, the ML emulator generates 8,000 years of RCM-resolution data that can then be used to find high resolution analogs for the low resolution extremes from ensemble boosting.}
    \label{fig:overview}
\end{figure}

As illustrated in \Cref{fig:overview}, our hybrid framework couples low-resolution ensemble boosting with high-resolution ML-generated data to efficiently downscale extreme weather events.
In this work, we first address whether the in-distribution extremes generated by the ML emulator (\Cref{fig:overview}, right part) reliably match target RCM dynamics. We use a machine learning emulator pre-trained on physics-based simulations (\Cref{sec:enscale}) to generate an 8,000-summer dataset, including extreme realizations.
Concurrently, ensemble boosting (\Cref{fig:overview}, left part) takes an extreme event and generates an ensemble of similar extreme events with a GCM (\Cref{sec:boost}).
Finally, the two parts of the framework are coupled via spatial average pooling and analog matching (\Cref{sec:analog-met}). For downscaling, each low-resolution boosted extreme is matched to a corresponding high-resolution field within the emulated dataset. In addition, we also show how analog matching can be used to identify alternative storylines and sample model variability by selecting extreme samples from other models without spatial field constraints.

\subsection{\textit{EnScale} setup}
\label{sec:enscale}
The \textit{EnScale} machine learning model was developed to emulate regional climate models (RCMs) given global climate model (GCM) inputs \citep{schillinger2026enscale}. It is a generative model that learns the conditional distribution of high-resolution RCM targets given GCM predictors, enabling the sampling of multiple realizations of high-resolution fields for a given GCM field. In its multivariate setup, it predicts temperature, precipitation, surface wind, and solar radiation jointly, spatially consistent across Central Europe. \textit{EnScale} splits the full GCM-RCM map into several physically-motivated steps: It first performs a ``coarse correction'', predicting a coarsened (average-pooled) version of the RCM data from GCM data. This step accounts for large-scale mismatches between GCM and RCM. The coarsened RCM fields are then iteratively refined through a sequence of super-resolution steps, each enhancing the spatial resolution further until the target resolution is recovered. Each stage is trained by minimizing the energy score, a strictly proper scoring rule that ensures \textit{EnScale}'s learned conditional distribution matches the target distribution.

In this paper, we use the version \textit{EnScale-t}, which explicitly models the temporal dependence of the RCM data through an autoregressive framework. Both \textit{EnScale} and \textit{EnScale-t} have been shown to accurately reproduce the multivariate conditional distribution. Compared to other state-of-the-art generative models, \textit{EnScale(-t)} performs competitively while requiring substantially lower computational costs \citep{rampal2026cordex}. Since we use only the temporal variant throughout, we refer to \textit{EnScale-t} simply as \textit{EnScale} in the following.

Here, we use eight RCM-GCM combinations spanning the years 1971-2099 (with training data between 1971-2030 and 2040-2089) at a daily resolution; an overview of the models is given in \Cref{tab:models}. The GCMs are from CMIP5 \citep{taylor2012overview}, and the RCMs are from CORDEX \citep{jacob2014euro}, derived from historical runs and RCP8.5. Although this emulator allows sampling from a specific target RCM, the shared representations learned during joint training mean that the generated samples from a specific RCM-GCM combination are not expected to perfectly reproduce the isolated characteristics of the original physical model. We introduce an adaptation to the standard \textit{EnScale-t} model. In its original configuration, temporal dependence is modeled at every spatial resolution. Here, for simplicity, we use a slightly adjusted version and restrict temporal autocorrelation modeling exclusively to the lowest spatial resolution (as suggested in the preprint \citep{schillinger2025enscale} of \citet{schillinger2026enscale}). Specifically, while the coarsened RCM data is modeled autoregressively, the super-resolution module is applied independently to each daily timestep without temporal conditioning. Across our evaluations, this modification yields no qualitative performance differences compared to the full baseline (\Cref{fig:combined-paris-plots}). 

\begin{table}[h!]
\centering
\small
\caption{RCM–GCM combinations in the emulator dataset. We enumerate all models according to the numbering displayed here. Numbers indicate model combinations that exist and were used both for training the machine learning model and in the data analysis presented here.}
\label{tab:models}

\begin{tabular}{lccc}
\hline
RCM / GCM & CNRM-CM5 & MIROC5 & MPI-ESM1-2-LR \\
\hline
ALADIN63    & model 1 &          & model 2    \\
CCLM4-8-17  & model 3 & model 4 & model 5   \\
REMO2015    &            & model 6 &               \\
RegCM4-6    & model 7 &          & model 8   \\
\hline
\end{tabular}
\end{table}

\paragraph{ML sampling and sample sizes}
Using \textit{EnScale}, we sample 8,000 emulator summers and compare them against a baseline of 8 ground-truth RCM-GCM combinations, which span 10 years of data each from 2030 to 2039 (totaling 80 summers). This means that, for each original RCM summer, we have 100 \textit{EnScale} samples. While the emulator is therefore expected to produce longer and more extreme events than are present in the original RCM data, the aim is to validate whether these generated extremes are physically plausible.

 \subsection{Ensemble boosting setup}
 \label{sec:boost}
 The ensemble boosting method \citep
 {gessner2021very} is a rare event algorithm that allows for a computationally efficient simulation of extreme weather events. Starting from a set of extreme events simulated in a large ensemble (``parent events''), it consists of adding round-off-sized perturbations to the atmospheric humidity field a few days or weeks before the peak of the parent event. In this way, a new ensemble of simulations is created, which quickly deviates substantially from their parent event.

 \begin{figure}[H]
    \centering
    \includegraphics[width=\linewidth]{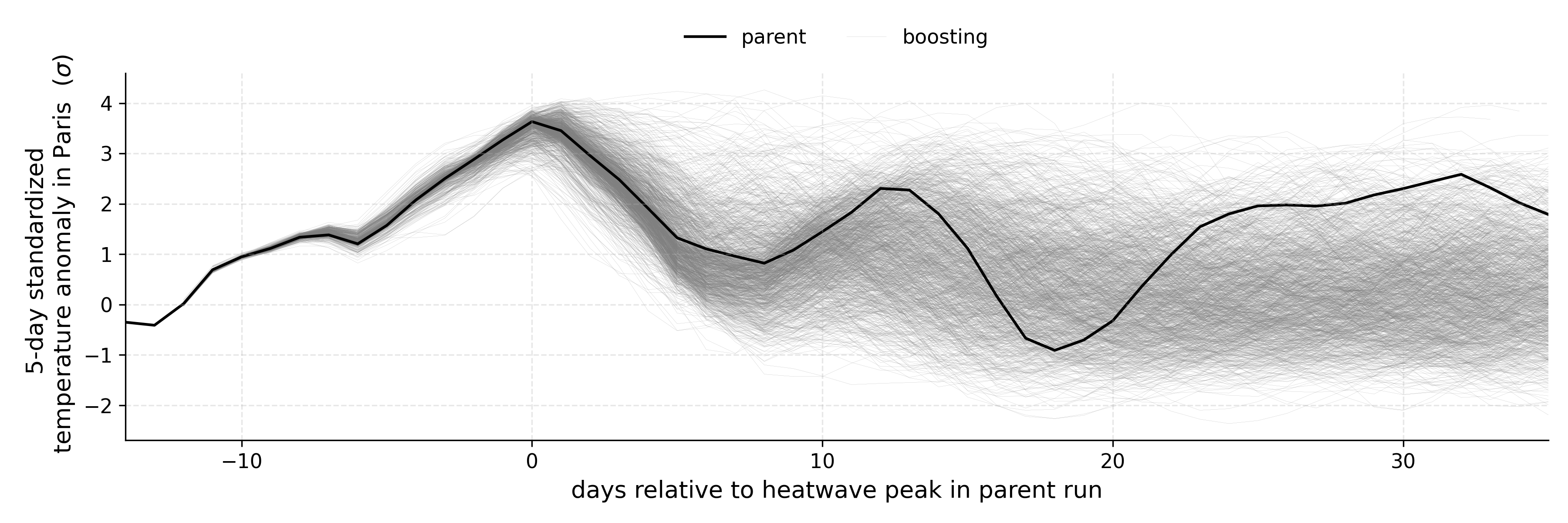}
    \caption{Ensemble boosting simulations for one extreme parent events. Shown are the temperature anomalies in Paris for the GCM's parent simulation (black) and the boosting ensemble (1200 simulations, gray).}
    \label{fig:boosting}
\end{figure}

 As an example, we use existing ensemble boosting simulations of heatwave storylines in the Greater Paris region, described in \citet{fischer_storylines_2023}. For these simulations, three parent heatwaves are selected based on the 5-day daily maximum temperature anomaly in Greater Paris in a CESM2 initial condition large ensemble for 2005--2035. We use simulations initialized 4--15\,days before the peak anomaly in the respective parent event, with 100 ensemble members per lead time. This yields $3\ \text{parent events} \times 12\ \text{initialization days} \times 100\ \text{ensemble members} = 3600\ \text{simulated heatwaves}$, each of which is run for 50\,days. The heatwaves occur in the model years 2016, 2028, and 2030 \citep{fischer_storylines_2023}.

\subsection{Downscaling via analogs}
\label{sec:analog-met}
In \Cref{sec:analogs}, we employ an analog approach to downscale the GCM ensemble-boosted extreme events to RCM resolution.
We choose the analog framework over direct ML-based downscaling to leverage the pre-trained emulator without the need or expense of retraining. Directly applying the trained emulator to boosted data would be challenging for several reasons: CESM2 was not part of the original training set, and the emulator’s required predictor variables and domain do not align with the saved output variables of the boosted dataset.

For analog-based downscaling, we use a two-step approach that first filters candidates by local Paris heatwave intensity using 2m temperature data and subsequently selects the most structurally similar event based on spatial pattern alignment. To construct the analog pool, we coarsen the 8,000 summers of \textit{EnScale} data to the GCM grid using conservative remapping via CDO \citep{schulzweida_2023_10020800}. Then, for a given Paris heatwave generated by the GCM, we identify all candidate 5-day heatwaves in the coarsened \textit{EnScale} dataset whose intensity falls within $\pm 0.1^\circ\text{C}$ of the GCM event; we describe the details in \Cref{sec:analogs}. From this pre-filtered subset, we select the event with the highest spatial similarity across the entire spatial field as the best analog. To assess spatial pattern alignment between the coarsened \textit{EnScale} field $x$ and the boosted GCM field $y$, we employ the Structural Similarity Index (SSIM; \citealp{wang2004image}) on 5-day temperature averages. Conceptually, SSIM aligns with object-based meteorological evaluation metrics such as the SAL (Structure, Amplitude, Location) index \citep{wernli2008sal}. SSIM evaluates spatial similarity by simultaneously comparing localized mean, contrast, and structural information:

\begin{equation}
\text{SSIM}(x, y) = \frac{(2\mu_x\mu_y + C_1)(2\sigma_{xy} + C_2)}{(\mu_x^2 + \mu_y^2 + C_1)(\sigma_x^2 + \sigma_y^2 + C_2)}
\end{equation}

where $\mu_x$ and $\mu_y$ denote the spatial means, $\sigma_x^2$ and $\sigma_y^2$ represent the spatial variances, and $\sigma_{xy}$ is the spatial covariance between the two fields. The terms $C_1 = (K_1L)^2$ and $C_2 = (K_2L)^2$ are stabilizing constants, where $L$ is the range of the temperature data and $K_1 =0.01, K_2 = 0.03$. The index ranges from $-1$ to $1$, where $1$ indicates perfectly identical spatial structures. Once the analog with the maximum SSIM is selected, we map it back to its original high-resolution \textit{EnScale} counterpart, yielding a physically plausible high-resolution analog for the GCM extreme event.

\subsection{Standardized Anomalies}
\label{sec:standardized}
To compare heatwaves across different models--especially those that fall outside the training RCM distributions--we compute standardized anomalies. Because \textit{EnScale} is trained jointly on all RCM-GCM combinations, individual model samples are influenced by the broader ensemble. Therefore, we standardize the data using the mean and variance computed across all models. For a given temperature $T_{t, y}$ on day $t$ of year $y$ (pre-smoothed with a 5-day running mean), the standardized anomaly $Z_{t, y}$ is computed as:
\begin{equation}
    Z_{t, y} = \frac{T_{t, y} - \mu_{t, y}}{\sigma_{t, y}}
\end{equation}
where the seasonal mean $\mu_{t, y}$ and standard deviation $\sigma_{t, y}$ are calculated over a rolling window of $\pm 2$ days and a reference climatology of $\pm 15$ years centered on the target date $(t, y)$:
\begin{equation}
    \mu_{t, y} = \frac{1}{N} \sum_{j=-15}^{15} \sum_{i=-2}^{2} T_{t+i, y+j}
\end{equation}
\begin{equation}
\label{eq:clim}
    \sigma_{t, y} = \sqrt{\frac{1}{N} \sum_{j=-15}^{15} \sum_{i=-2}^{2} (T_{t+i, y+j} - \mu_{t, y})^2}
\end{equation}
The total number of RCM days used to compute the climatology is $31 \mathrm{years} \times 5 \mathrm{days} = 155$ samples.

For the ensemble boosting, the same approach is used to calculate standardized anomalies. As climatology, we use the 100-member CESM2 large ensemble running for the period 1850--2100 under historical and SSP370 scenarios, respectively \citep{danabasoglu_cesm2_2020,rodgers_ubiquity_2021}. This allows for the construction of centered climatologies for all ensemble boosting simulations.

\subsection{Weighted least squares}
\label{wls}
To evaluate heatwave persistence and physicality, we use a unified Weighted Least Squares (WLS) regression framework. Intuitively, this method fits a linear trend to each model class simultaneously, and we use it to perform hypothesis testing to determine whether the ML emulator \textit{EnScale} exhibits statistically significant slope departures from physics-based baselines.

For any target variable $Y$ evaluated against its driver $x$, we estimate the single interaction model:
\begin{equation}
Y = \alpha_0 + \beta_1 x + \alpha_{\text{RCM}} I_{\text{RCM}} + \alpha_{\text{GCM}} I_{\text{GCM}} + \beta_{\text{int,RCM}} (x \times I_{\text{RCM}}) + \beta_{\text{int,GCM}} (x \times I_{\text{GCM}}) + \varepsilon
\end{equation}
where $I_{\text{RCM}}$ and $I_{\text{GCM}}$ are binary indicators ($1$ if the data comes from RCM or GCM respectively, $0$ otherwise).

Using \textit{EnScale} as the baseline, $\beta_1$ isolates the emulator's slope, while the interaction terms $\beta_{\text{int,RCM}}$ and $\beta_{\text{int,GCM}}$ directly measure the slope differences of the RCM and GCM relative to \textit{EnScale}. Thus, the null hypotheses are that these interaction terms are $\beta_{\text{int,RCM}}=0$ and $\beta_{\text{int,GCM}}=0$, respectively. A two-tailed $t$-test within a WLS ANCOVA framework (\citep[\eg][]{barrett2011computations}) evaluates whether the slope differences are statistically significant. To ensure robustness, each bin is weighted by its sample count $N$, giving more influence to densely populated bins and preventing rare events from skewing the trend line.

To evaluate heatwave persistence, we map this framework directly to an exponential decay model $P(d) \propto \exp(\beta d)$ by setting $Y = \ln P(d)$ (the log-probability of an event lasting $d$ days) and $x = d$ (event duration in days). In this setup, $\beta_1$ measures the emulator's duration decay rate, from which we derive the daily continuation probability $p_{\text{cont}} = \exp(\beta_1)$ and the atmospheric memory scale $\tau = -1/\beta_1$.

\section{Validation of \textit{EnScale} heatwaves}
\label{sec:validation}
To use \textit{EnScale} data for downscaling ensemble boosted heatwaves, we first validate heatwaves in \textit{EnScale} using a twofold approach. First, for in-distribution heatwaves, we verify that RCM-simulated heatwaves in Paris closely match ML-generated heatwaves in magnitude, temporal alignment, and spatial fields. We then extend this analysis to out-of-distribution heatwaves.

\subsection{In-distribution heatwaves}
\begin{figure}[H]
    \centering
    \includegraphics[width=\linewidth]{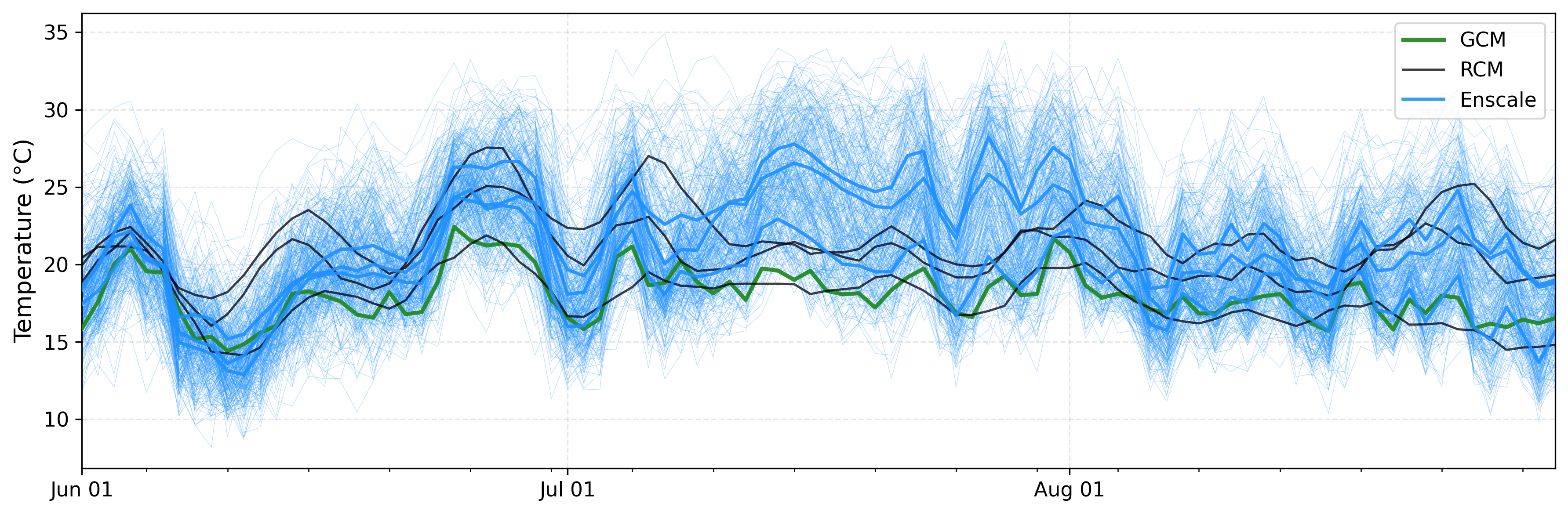}
    \caption{Average daily temperature time series in the MPI model and corresponding physical RCMs and \textit{EnScale} for the summer 2033. Thick lines show mean values for the different model classes (one per RCM), while thin lines display individual realizations.}
    \label{fig:temporal-1model}
\end{figure}

To ensure that emulated heatwaves do not emerge spuriously, we first verify that the temperature time series of both RCMs and \textit{EnScale} align with the temporal progression of their driving GCMs, as visualized in \Cref{fig:temporal-1model} (for an overview separated by GCM, see \Cref{fig:temporal}). Overall, the correlation between GCMs and RCMs and emulated RCMs, respectively, is high. The lower individual member correlation of \textit{EnScale} and the GCMs ($0.61$--$0.74$ for the year 2033) paired with a higher ensemble-mean correlation ($0.82$--$0.93$ for the year 2033; \Cref{tab:gcm_correlations}) demonstrates that \textit{EnScale} successfully separates fluctuations from the large-scale atmospheric variability imposed by the driving GCM. While 2033 is selected here as a representative year, this behavior is consistent and robust across the other years as well. 

\begin{figure}[H]
    \centering
    \includegraphics[width=\linewidth]{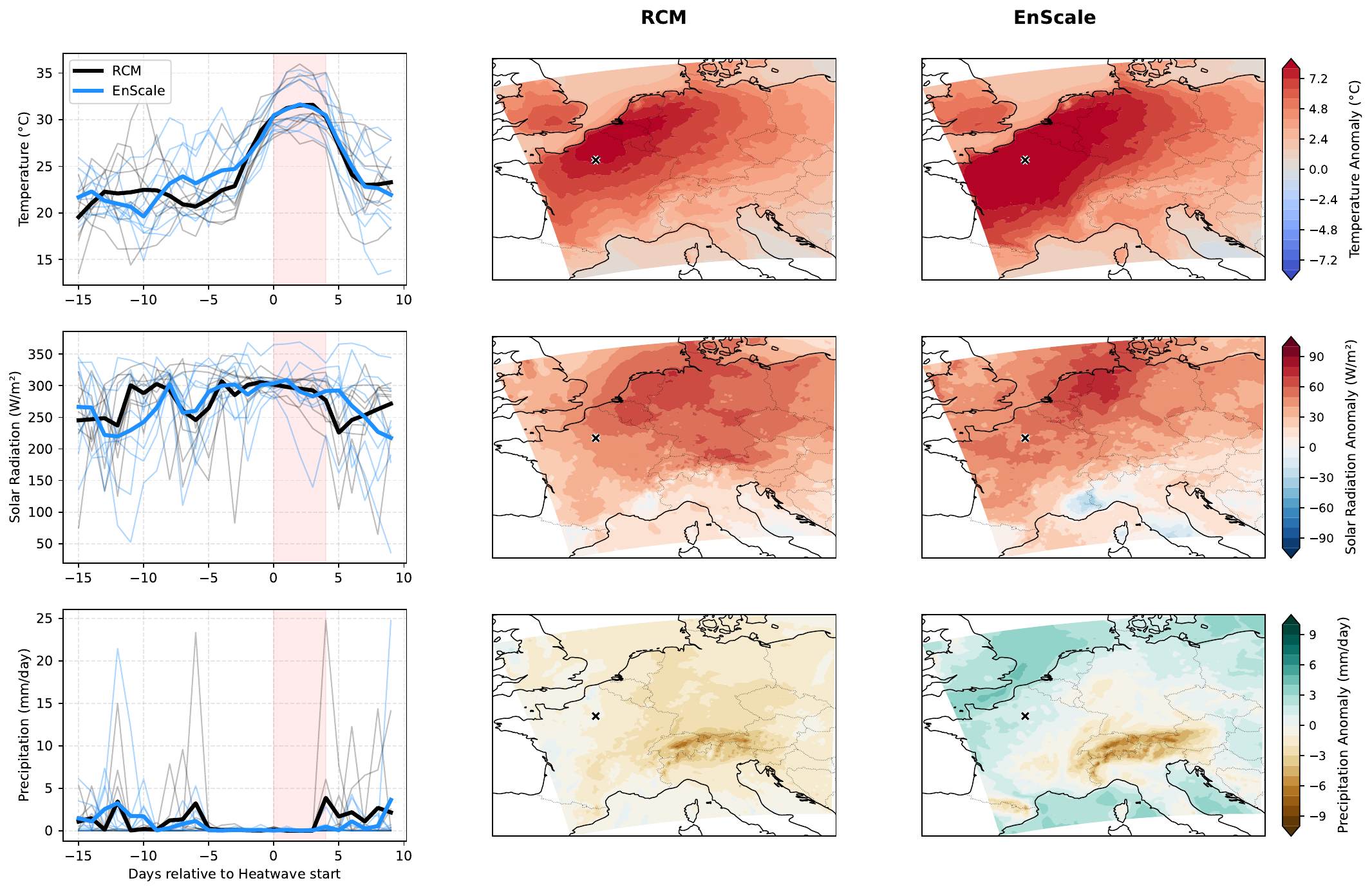}
    \caption{Temporal and spatial alignment of analog heatwaves in RCM and \textit{EnScale}. 
    Comparison of the 8 hottest 5-day heatwaves in Paris (2030--2039) across physical models (RCM-GCM combinations, black) and their closest \textit{EnScale} analogs (blue), selected by minimizing the mean squared error (MSE) of Paris temperature over the 5-day heatwave period (red shading). 
    Left panels: Averaged 15-day temporal trajectories centered around the heatwave for temperature (top), solar radiation (middle), and precipitation (bottom). 
    Right panels: Corresponding composite spatial anomaly fields (deviations from climatology; \Cref{eq:clim}) averaged across all 8 heatwave events during the 5-day peak period for the original RCM (left sub-column) and \textit{EnScale} (right sub-column). 
    The black 'x' denotes the location of Paris.}
    \label{fig:spatial-best-fit}
\end{figure}

Building on this broad temporal consistency, we find that \textit{EnScale} can reproduce the intensity of 5-day RCM heatwaves. Selecting the hottest 2030–2039 heatwaves in Paris per physical RCM (8 in total), ML-generated analogs of \textit{EnScale} closely mirror individual RCM events at both local and spatial scales (\Cref{fig:spatial-best-fit}). To formally test this alignment, we perform a permutation test measuring the 5-day temperature trajectory RMSE relative to the ensemble reference. Sampling random groups of 8 events from the \textit{EnScale} distribution ($N=10{,}000$) reveals that the 8 original RCM heatwaves are statistically indistinguishable from random \textit{EnScale} samples (for detailed results, see \Cref{fig:permutation-test}). We then find the closest matches for each of the 8 hottest RCM heatwaves (\Cref{fig:spatial-best-fit}). Selecting analogs based on minimizing the mean squared error for temperature across the hottest 5 days, we find that the corresponding solar radiation is high (middle row), and precipitation is zero or close to zero before and at the beginning of the heatwave (last row), suggesting similar physical properties of the heatwaves. 

Comparing the composite spatial fields averaged across all 8 heatwaves yields a high spatial correlation (0.988) and SSIM (0.980), confirming that \textit{EnScale} accurately reconstructs the two-dimensional spatial structure and correctly centers the localized temperature maximum over Paris. Similarly, solar radiation is very high in Paris and the surrounding areas; precipitation is very low in Paris. However, surrounding areas show more rain in \textit{EnScale} compared to the original RCM. Crucially, analogs were identified based on the temporal alignment of temperatures at the Paris grid point only, without spatial field constraints during selection. Therefore, the spatial fields and other variables are not expected to be identical.

\begin{figure}[H]
        \centering        
        \includegraphics[width=\linewidth]{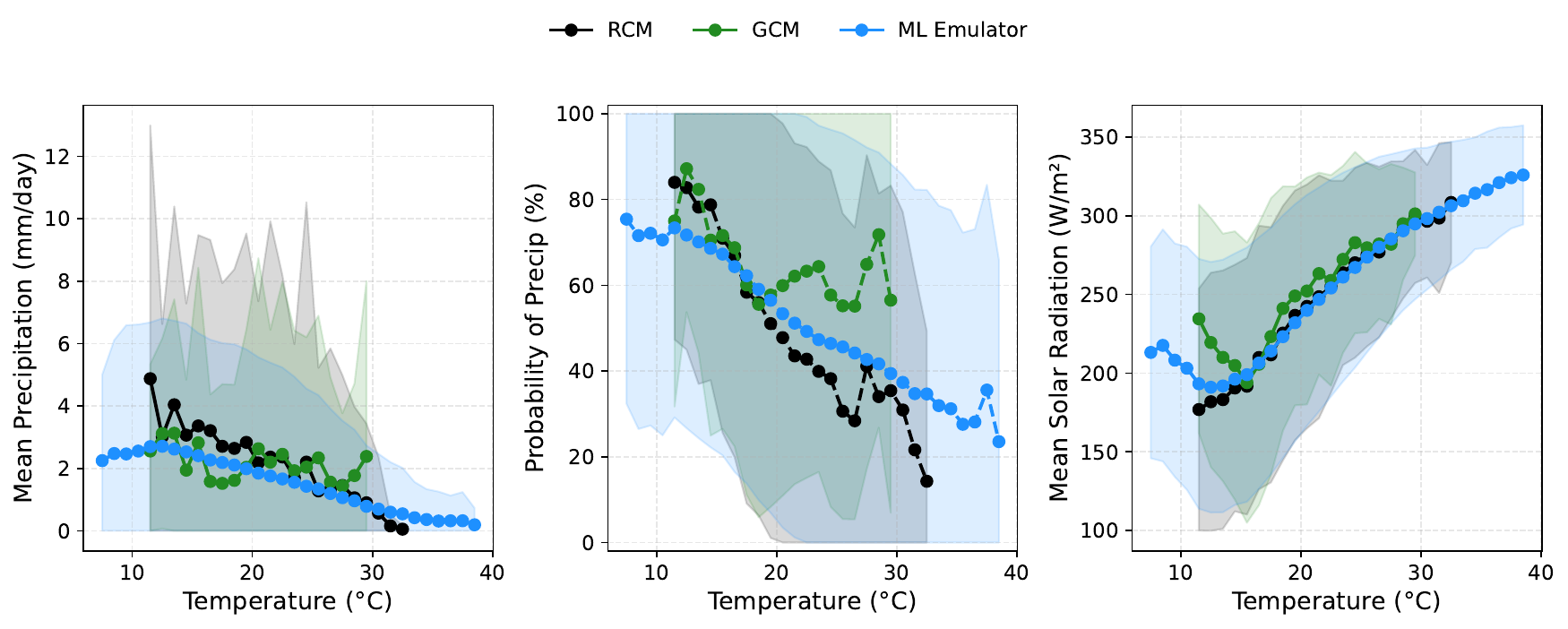}
    \caption{Comparison of temperature sensitivities at the Paris grid point across the whole temperature ranges, binned in $1^\circ\text{C}$ increments, for precipitation amount (left), precipitation probability (center), and mean solar radiation (right) across \textit{EnScale}, RCM, and GCM. Shaded regions denote $\pm1$ standard deviation of daily values within each bin.}
    \label{fig:physicality}
\end{figure}

To further evaluate intervariable consistency, we analyze whether \textit{EnScale} variables behave as physically expected across the range of temperatures. \Cref{fig:physicality} reveals distinct patterns of climate parameter sensitivities across GCMs, RCMs, and \textit{EnScale}. For the amount and probability of precipitation $>0.1 \text{mm/day}$, \textit{EnScale} exhibits intermediate temperature sensitivities that fall between the physical RCMs and GCMs (for a figure separated by RCM, we refer to \Cref{fig:sep-by-rcm}). During the hottest days, mean precipitation sinks to almost zero in both RCM and \textit{EnScale} data. Specifically, for temperatures above $29^\circ\text{C}$, precipitation is usually less than 1 mm/day. However, overall precipitation is slightly overestimated in the ML samples for hotter temperatures. However, because our selection relies strictly on temperature thresholds, some identified heatwaves may include intense late-day convective storms \citep{sauter2023compounding}, introducing rain on peak temperature days and making precipitation a noisy evaluation metric. For mean solar radiation, a WLS regression reveals that all three model classes demonstrate statistically indistinguishable sensitivities (\Cref{tab:climate_sensitivities}), increasing at approximately $6.59$ to $7.26~\text{W/m}^2$ per $^\circ\text{C}$. Although solar radiation amounts vary between RCM, GCM and \textit{EnScale} below $15 ^\circ\text{C}$, they increase at temperatures above $15 ^\circ\text{C}$ across all three model classes, exhibiting similar slopes and absolute values.

\subsection{Out-of-distribution heatwaves}
\label{out-of-dist}
Having established that \textit{EnScale} reproduces in-distribution heatwaves, including realistic multivariate aspects and the temporal dynamics of the GCM, we now evaluate \textit{EnScale}'s capacity to emulate out-of-distribution extreme events. Using anomaly datasets spanning the same 80 years of RCM simulations and 8,000 years of \textit{EnScale} emulations, we evaluate whether \textit{EnScale} can generate unprecedented extremes while maintaining physical realism. To better compare heatwaves of different RCM-GCM combinations, we compute standardized anomalies as explained in \Cref{sec:standardized}. 

\begin{figure}[H]
        \centering
        \includegraphics[width=\linewidth]{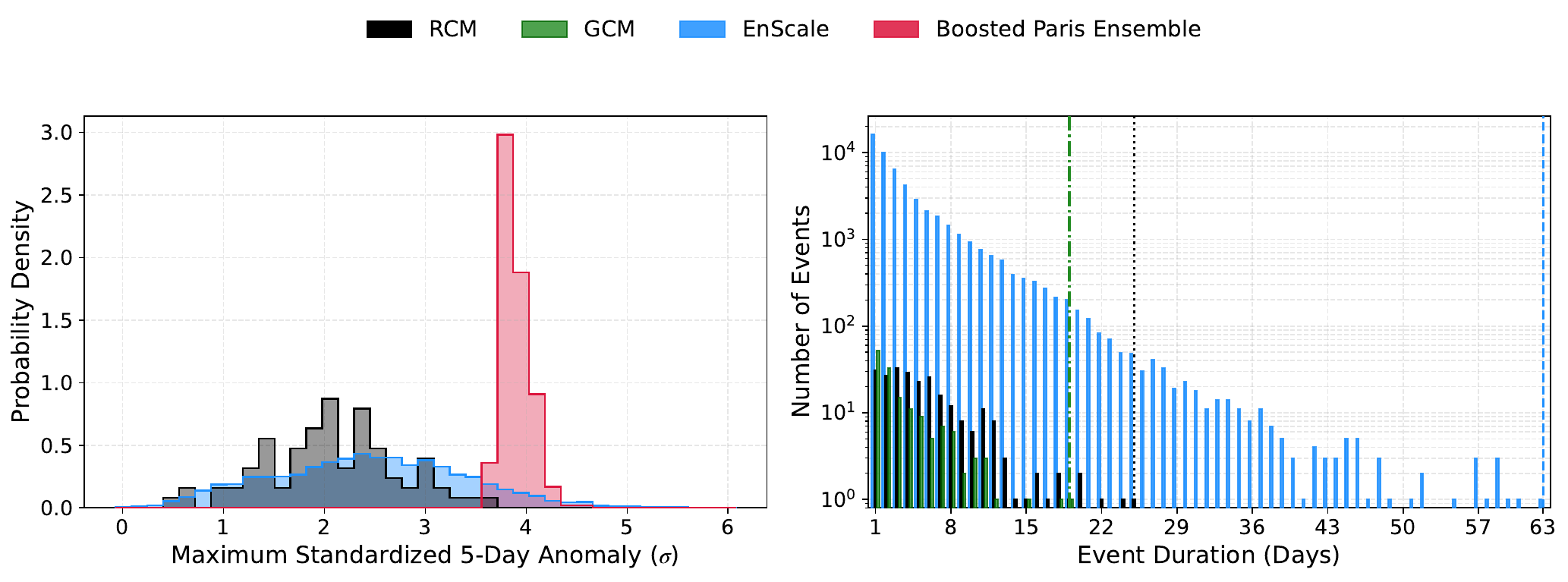}
    \caption{a) Histograms of maximum standardized 5-day anomalies in Paris of each summer in boosted data, GCMs, original RCMs and \textit{EnScale} ensemble members. b) Frequency of $1\sigma$ extreme events as a function of event duration. Vertical dashed lines mark the duration of maximum frequency for each model class.}
    \label{fig:combined-ood-extremes}
\end{figure}

We first evaluate the distribution of 5-day standardized temperature anomalies (in $\sigma$). As shown in \Cref{fig:combined-ood-extremes} (left), \textit{EnScale} covers the full range of anomaly distributions; however, its expanded 8,000-year sample size generates extreme intensities beyond RCM bounds and into the range of samples from ensemble boosting (\Cref{sec:boost}). While the maximum RCM anomaly is $3.56\sigma$, \textit{EnScale} produces events of up to $6.08\sigma$, compared to $4.64\sigma$ in ensemble boosting. The frequency of these very extreme tail events drops with increasing magnitude. Crucially, these high-magnitude anomalies are strongly RCM-dependent rather than GCM-driven: over 80\% originate from \textit{RegCM4-6} and \textit{REMO2015}. Conversely, \textit{CCLM4-8-17} produces almost no events in these upper bins, while \textit{ALADIN63} displays intermediate counts, mainly in the $[4.6,5.2\sigma)$ range (\Cref{tab:idx_threshold_breakdown}). This RCM dominance over driving GCMs aligns with established literature: while winter climate is dictated by large-scale synoptic advection, summer heat extremes are governed by local land-surface feedbacks and convective parameterizations that differ across RCMs \citep{fischer2007contribution}.

To test whether \textit{EnScale} alters heatwave duration characteristics relative to the physics-based baselines, we evaluate whether the model classes share the same daily probability decay slope ($\beta$) using WLS regression. The distribution of heatwave lengths is visualized in \Cref{fig:combined-ood-extremes} (right) and shows a decay in the number of events with increasing heatwave length. The longest GCM, RCM, and \textit{EnScale}  $1\sigma$ heatwaves are 18, 25, and 63, respectively.   
\textit{EnScale} accurately captures the exponential temporal decay of multi-day $1\sigma$ heatwave events, remaining consistent with both the raw RCM reference and the driving GCMs. The log-linear probability decay yields a daily slope of $\beta=-0.2773$ for \textit{EnScale}, corresponding to a 75.8\% daily continuation probability ($e^\beta$). This translates to an atmospheric memory scale ($-\frac{1}{\beta}$) of $\tau=3.61$ days. While \textit{EnScale} exhibits slightly higher temporal persistence than the RCM ($\tau=2.98$ days) and GCM ($\tau =3.29$ days), the regression interaction terms confirm that these decay rate differences are statistically insignificant (\Cref{tab:prob_slope_results}), suggesting \textit{EnScale} retains physically consistent heatwaves.

Looking at the most extreme heatwaves per RCM-GCM combination reveals spatial similarities across all models (\Cref{fig:spatial-max}). The maximum values again depend on the RCMs, with ALADIN, RegCM, and REMO generating much higher temperatures than CCLM (5-day mean daily temperature in Paris $>37^\circ\text{C}$ vs. $<35^\circ\text{C}$.) Most very strong heatwaves in \textit{EnScale} ($>4.6\sigma$) occur in June; see \Cref{fig:temporal-distribution}.

\section{Applications}
\subsection{Analog-based downscaling of ensemble boosting events}
\label{sec:analogs}
Analog-based downscaling of extremes relies on the assumption that the target search space contains high-resolution realizations that are physically plausible and representative of rare tail events. The previous section establishes this foundation by demonstrating that \textit{EnScale} preserves spatial and temporal structures, multivariate physical relationships, and extreme-value decay rates under baseline conditions. We therefore conclude that \textit{EnScale}'s sample space ($N=8,000$ summers) functions as a physically credible library of high-resolution climate states. As it also spans the range of extremes observed in ensemble boosting (\Cref{fig:combined-ood-extremes} (left)), we use it to downscale ensemble-boosted GCM extremes via analog matching, as described in \Cref{sec:analog-met}. Specifically, we first select days when the temperatures in Paris are similar to those found in boosting ($ 35^\circ\text{C} \pm 0.1^\circ\text{C}$) and then search for high-resolution spatial analogs corresponding to the peak heatwave periods in the boosted GCM time series.

\begin{figure}[H]
    \centering
    \includegraphics[width=0.9\linewidth]{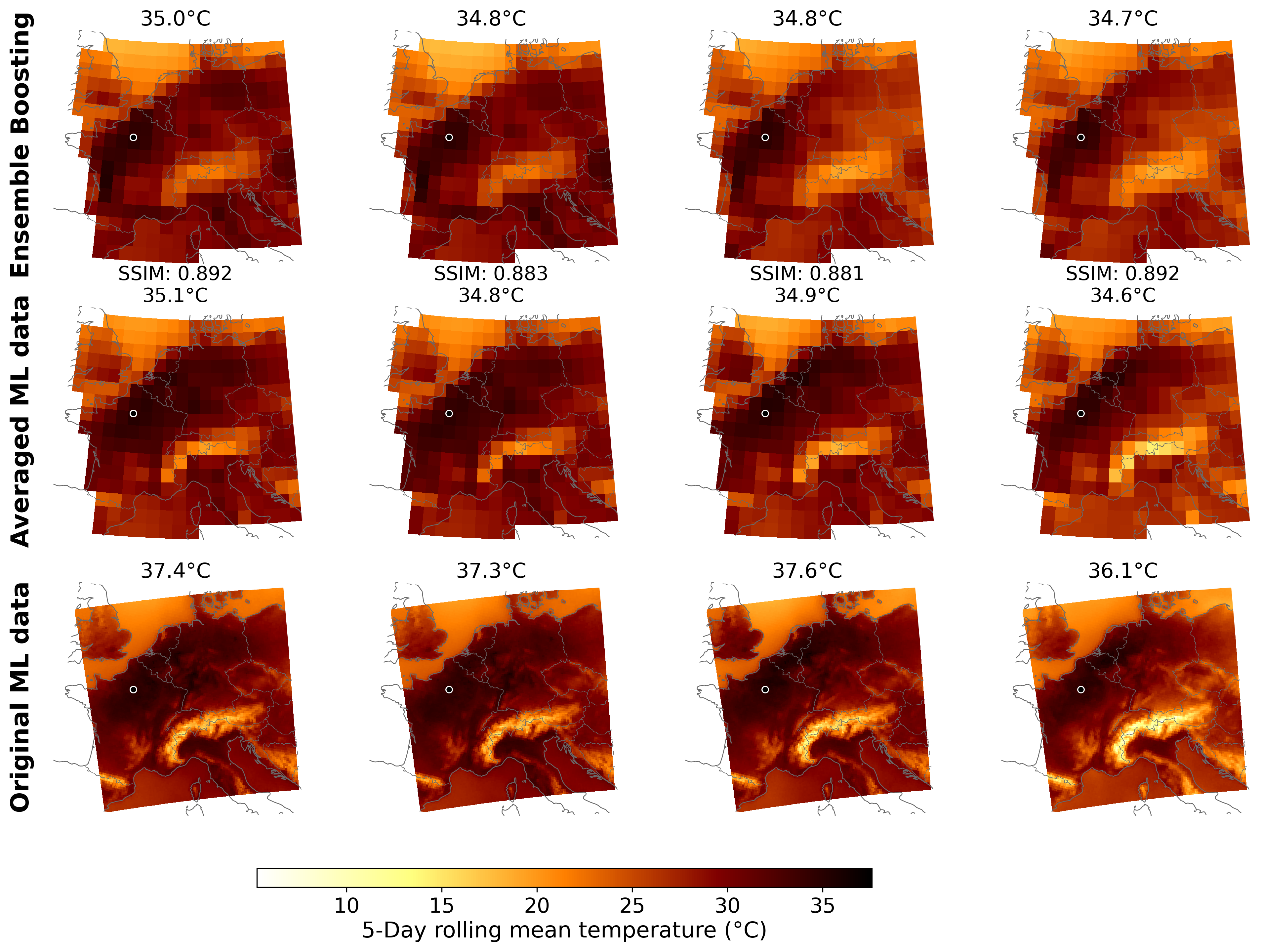}
    \caption{The four hottest events from ensemble boosting (top row), alongside their corresponding \textit{EnScale} analogs at coarse average-pooled resolution (middle row) and native \textit{EnScale} resolution (bottom row). Values above each panel show the temperature at the grid point closest to Paris. Annotations above the middle row report the Structural Similarity Index between the original boosted events and the coarsened \textit{EnScale} fields. The black 'o' denotes the location of Paris.}
    \label{fig:boosting-analogs}
\end{figure}

\Cref{fig:boosting-analogs} illustrates the heatwaves and the corresponding downscaling results for the four hottest events (in terms of $\sigma$) identified in the ensemble boosting simulations. For each coarse GCM boosted event, the \textit{EnScale} search space yields matching high-resolution analogs (SSIM $\approx 0.88$) that reconstruct fine-scale regional features, such as localized heat anomalies around Paris and topographic attenuation over the Alps, while adhering to the synoptic-scale circulation patterns of the original GCM data. The (average-pooled) RCM samples exhibit a cold bias over the Alps compared to the GCM data, consistent with \eg \citet{rudisill2024atmospheric}. Across all four cases, the high-resolution analogs yield higher temperatures than their low-resolution counterparts, ranging from $36.1^\circ\text{C}$ to $37.6^\circ\text{C}$, underscoring the need for spatial downscaling to capture true localized severity.

To evaluate whether the spatial error introduced by analogue matching is acceptable, we compare the downscaling mismatch against RCM multi-model variability. Spatially, the multi-model variability across the most extreme heatwaves per RCM-GCM combination (those from \Cref{fig:spatial-best-fit}) averages $1.80\,^{\circ}\text{C}$. While our two-step approach constrains the temperature error in Paris to $<0.1\,^{\circ}\text{C}$ before matching spatial patterns, the average root-mean-square error (RMSE) of the eight most extreme events averages $2.21\,^{\circ}\text{C}$ (the difference between row 1 and 2 in \Cref{fig:boosting-analogs}). As shown in \Cref{fig:var_vs_rmse}, downscaling errors exceed RCM variability primarily over sea areas and mountainous regions such as the Alps. This mismatch is not surprising since coarse climate models smooth out mountain ranges \citep{gutowski2020ongoing} and coastlines \citep{di2013potential}, whereas detailed regional models capture these local differences better. Over land, particularly across northern France and the broader Paris region, the mismatch remains lower than or comparable to the RCM spread. Within $\pm 3$ GCM grid cells around Paris ($\approx 600 \times 730\text{ km}$), EnScale achieves an RMSE of $1.43\,^{\circ}\text{C}$ against an RCM spread of $1.54\,^{\circ}\text{C}$. Because the spatial downscaling error stays within the range of RCM ensemble uncertainty over the larger Paris region, the analog-matching approach proves fit for purpose for localized climate impact modeling.

\subsection{Sampling model variability}
\begin{figure}[H]
    \centering
    \includegraphics[width=0.9\linewidth]{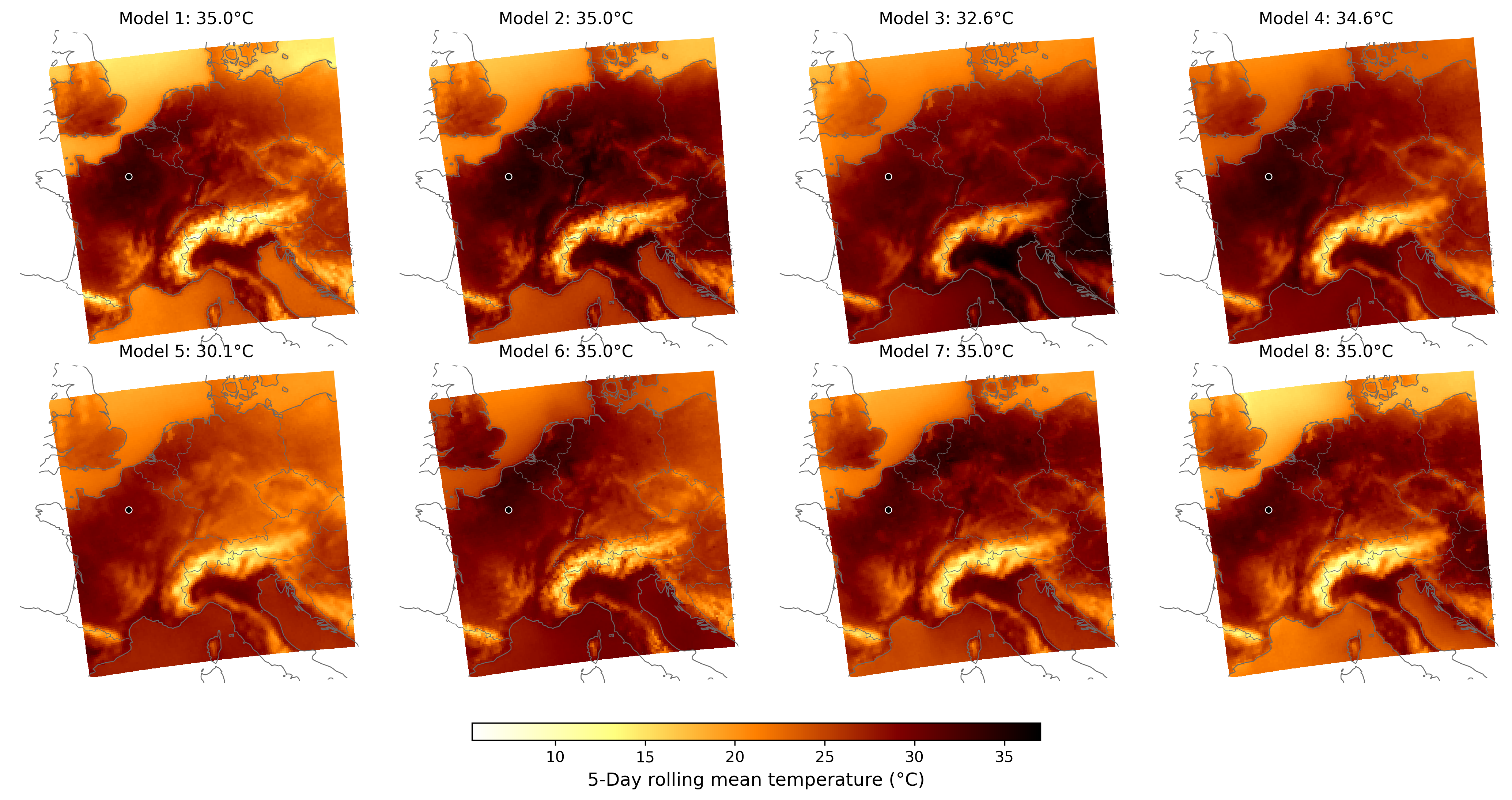}
    \caption{Spatial fields of $35^\circ\text{C}$ extreme events across \textit{EnScale} climate models. For each RCM-GCM combination, the panel shows the event where the grid point nearest Paris (indicated by a black 'o') most closely matches $35^\circ\text{C}$ with temperature at this grid point indicated above each subplot.}
    \label{fig:storylines}
\end{figure}

Beyond downscaling boosted extremes, \textit{EnScale} can also sample alternative heatwave storylines. By selecting events in the emulator data that are closest to the most extreme boosted event with a peak temperature of $35^\circ\text{C}$ over Paris, without constraining the remaining spatial field, we sample across regional model realizations to explore model uncertainty. The results are visualized in \Cref{fig:storylines} and highlight spatial differences: while some samples yield broad continental heatwaves (such as model 2), others feature more spatially constrained patterns (such as model 1). Notably, all global climate models downscaled with CCLM4-8-17 (models 3, 4 and 5) yield consistently lower temperatures for Paris that fail to reach $35^\circ\text{C}$. 

Ultimately, storyline sampling with multi-model ML emulators provides a practical way to explore climate risks beyond a single model simulation. By combining boosted GCM events with high-resolution RCM analogs, the framework shows that extreme heatwaves can unfold in different ways, from broad continental heatwaves to localized hot spots. Capturing these different spatial patterns and model-specific differences is essential for realistic uncertainty and risk assessment \citep{tebaldi2007use}. Generative machine learning models trained on a multi-model ensemble provide a computationally efficient way to sample model uncertainty and physically realistic extremes beyond what is possible with standard climate model runs.

\section{Discussion and Conclusion}
    While physical ensemble boosting enables targeted sampling of extreme climate events in coarse GCMs, high-resolution dynamical downscaling remains computationally unfeasible across large ensemble sets. Conversely, traditional statistical downscaling often fails due to a lack of historical analogs for unprecedented extremes. Our hybrid methodology overcomes these limitations by combining physical ensemble boosting with ML-driven analog downscaling across several emulated RCMs.

By applying the temporally consistent ML emulator \textit{EnScale}, pre-trained on physics-based GCM–RCM simulations, we generate a high-resolution dataset encompassing 8,000 simulated summers across 8 RCMs. We show that the ML samples span the range of extremes from ensemble boosting and are physically consistent. 
Our approach is applicable to a variety of extremes. For instance, it could easily be expanded to cold spells, and by working with two variables, renewable energy droughts could be analyzed \citep{bloin2025climate}. Application to new cases requires validating the physicality of the ML model for the specific task. Analog matching can further align with our two core objectives: downscaling to introduce spatial detail and identifying alternative storylines by selecting samples based on local events without spatial field constraints. Furthermore, analog matching is flexible with respect to the similarity metric used and can, for instance, also be performed on standardized anomalies ($Z$-scores) rather than absolute temperature fields, controlling for model biases or matching extremes across differing background climate states.
Our hybrid setup does not only increase spatial resolution to more impact-relevant scales. It also offers major efficiency gains by eliminating dynamical RCM runs, effectively capturing internal variability and, by emulating several regional models simultaneously, model uncertainty. Because the emulator is trained on GCM-RCM pairs, the samples also include GCM model uncertainty. 

Despite our promising results, several limitations offer directions for future work. A primary constraint is that the emulator datasets are not aligned with the ensemble-boosted data; future efforts could address this by leveraging regional CESM2 data for machine learning training. Additionally, the limited sample size of the original RCM data makes robust significance testing and cross-method comparisons difficult, as long return periods and very extreme events fall outside the range of what is captured by the original RCM data. Finally, it remains unclear whether the very extreme ML events that go beyond events observed in ensemble boosting ($>4.6\sigma$) are statistical artifacts or plausible scenarios. These extreme values may stem from our normalization, or because we can sample 10 full summers rather than just a few individual events, we may be capturing extreme scenarios with different spatial fields than those found in boosting.

\subsection*{Acknowledgments}
Nina Effenberger, Maybritt Schillinger, Maxim Samarin, and Reto Knutti are part of SPEED2ZERO, a Joint Initiative co-financed by the ETH Board.
\bibliographystyle{plainnat}
\bibliography{sample}
\newpage
\begin{appendices}
\section{Supplementary material}
\counterwithin{figure}{section}
\counterwithin{table}{section}
\setcounter{figure}{0}
\setcounter{table}{0}

\begin{figure}[htbp]
    \centering
    \includegraphics[width=\linewidth]{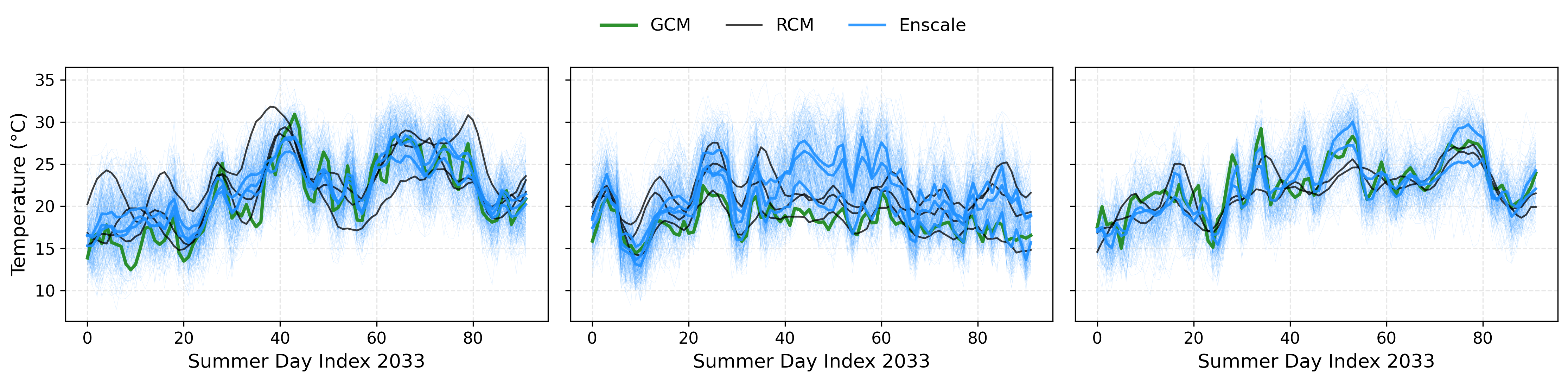}
    
    \caption{Average daily temperature time series in the MPI model, corresponding physical RCMs, and \texttt{EnScale} for summer 2033. Thick lines show mean values for each model class, while thin blue lines display individual \texttt{EnScale} realizations. Subpanels are separated by their driving GCM (from left to right: CNRM-CM5, MPI-ESM1-2-LR, and MIROC5).}
    \label{fig:temporal}
\end{figure}
\begin{table}[H]
\centering
\caption{Temporal Pearson correlation coefficients of daily mean temperature at the Paris grid point between downscaled models and their parent GCM in summer 2033.}
\label{tab:gcm_correlations}
\begin{tabular}{lccc}
\hline
 &{CNRM-CM5} &{MPI-ESM1-2-LR} &{MIROC5} \\ \hline
RCM vs GCM & 0.8497 & 0.7604 & 0.8119 \\
Enscale Members vs GCM (Avg) & 0.7370 & 0.6096 & 0.7075 \\
Enscale Ensemble Mean vs GCM & 0.9339 & 0.8187 & 0.8902 \\ \hline
\end{tabular}
\end{table}

\begin{table}[htbp]
\centering
\caption{Temperature sensitivity of hydrological parameters and surface solar radiation across model classes at the grid point closest to Paris.}
\label{tab:climate_sensitivities}
\begin{tabular}{lcccccc}
\toprule
& \multicolumn{2}{c}{\textbf{Precipitation Amount}} & \multicolumn{2}{c}{\textbf{Probability of Precip.}} & \multicolumn{2}{c}{\textbf{Solar Radiation}} \\
\cmidrule(lr){2-3} \cmidrule(lr){4-5} \cmidrule(lr){6-7}
\textbf{Model} & \textbf{Slope}\footnotemark[1] & \textbf{$p$-value}\footnotemark[2] & \textbf{Slope}\footnotemark[3] & \textbf{$p$-value}\footnotemark[2] & \textbf{Slope}\footnotemark[4] & \textbf{$p$-value}\footnotemark[2] \\
\midrule
ML Emulator & $-0.1094$ & -  & $-1.9567$ & - & $+6.5894$ &-\\
RCM         & $-0.1646$ & 0.00169   & $-3.1648$ & 0.01798   & $+7.2609$ & 0.61598 \\
GCM         & $-0.0144$ & 0.00167   & $-0.8457$ & 0.19794   & $+6.8547$ & 0.90798 \\
\bottomrule
\end{tabular}
\begin{flushleft}
\small
\footnotemark[1] mm/day per $^\circ$C \\
\footnotemark[2] $p$-value for slope difference relative to the reference ML Emulator \\
\footnotemark[3] \% per $^\circ$C \\
\footnotemark[4] W/m$^2$ per $^\circ$C
\end{flushleft}
\end{table}

\begin{table}[htbp]
\centering
\caption{Distribution of extreme daily temperature anomaly events at the Paris grid point during summer 2033 across $0.2\sigma$ threshold bins. Standardized anomalies ($\sigma$) are calculated for daily mean temperature relative to the summer reference climatology. Columns denote individual downscaling models (\texttt{model 1}--\texttt{model 8}); models 3 and 5 contained zero events exceeding $4.6\sigma$ and are omitted. Percentages in the final row represent the relative share of total extreme heat events contributed by each run.}\label{tab:idx_threshold_breakdown}
\begin{tabular}{lcccccc}
\toprule
 & model 1 & model 2 & model 4 & model 6 & model 7 & model 8 \\
Threshold Bin ($\sigma$) &  &  &  &  &  &  \\
\midrule
{[4.6, 4.8)} & 3 & 3 & 3 & 16 & 8 & 10 \\
{[4.8, 5.0)} & 0 & 1 & 0 & 7 & 6 & 6 \\
{[5.0, 5.2)} & 2 & 2 & 0 & 1 & 4 & 2 \\
{[5.2, 5.4)} & 0 & 0 & 0 & 5 & 4 & 1 \\
{[5.4, 5.6)} & 1 & 0 & 0 & 2 & 1 & 1 \\
{[5.6, 5.8)} & 0 & 1 & 0 & 1 & 1 & 0 \\
{[5.8, 6.0)} & 0 & 0 & 0 & 1 & 0 & 0 \\
{[6.0, 6.2]} & 0 & 0 & 0 & 1 & 0 & 0 \\
Relative Share (\%) & 6.32\% & 7.37\% & 3.16\% & 35.79\% & 26.32\% & 21.05\% \\
\bottomrule
\end{tabular}
\end{table}

\begin{figure}[H]
    \centering
        \includegraphics[width=\linewidth]{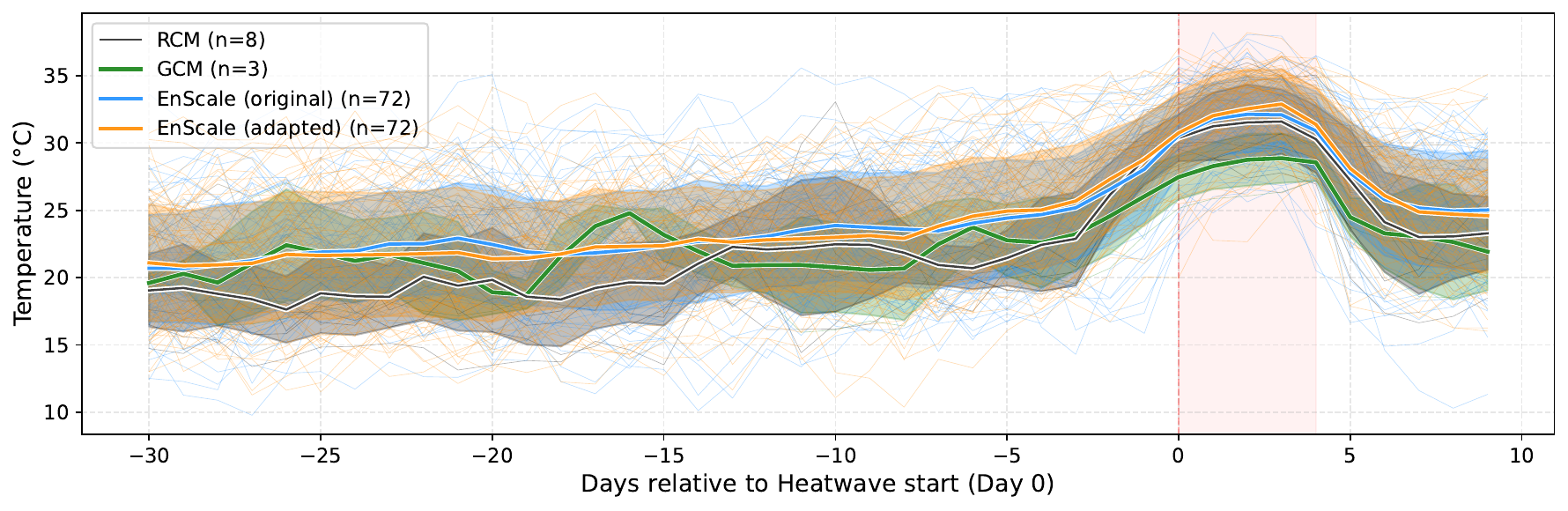}
   \caption{Temporal evolution of most extreme heatwaves at the Paris grid cell during 2030--2039. The red shaded background highlights the hottest 5-day average temperature window, along with the 30 preceding and 5 subsequent days. Thick lines display the ensemble mean daily temperature for each model class (GCM, RCM,  \texttt{EnScale} and \texttt{EnScale-t}), while thin lines depict individual ensemble realizations. Overall, the mean heatwave profile generated by \texttt{EnScale-t} closely aligns with that of the physical RCMs.}
    \label{fig:combined-paris-plots}
\end{figure}

\begin{figure}[H]
    \centering
    \includegraphics[width=\linewidth]{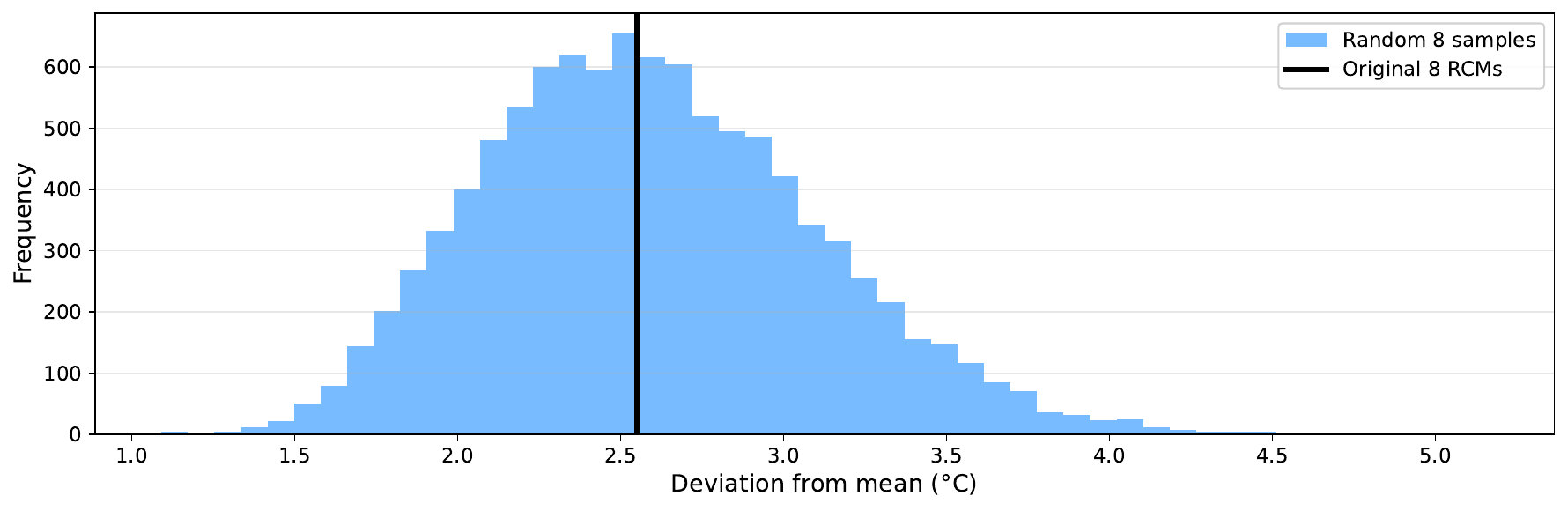}
    \caption{Histogram of permutation test assessing the similarity of heatwave profiles between physical RCMs and the ML emulator at the Paris grid point. The histogram shows the distribution of average RMSE ($^\circ\text{C}$) relative to the mean ML emulator heatwave, computed from 10,000 random samples of 8 ML realizations. The vertical black line indicates the observed mean RMSE across the 8 physical RCMs, demonstrating that the RCM heatwave deviation is statistically indistinguishable from random emulator sampling.} 
    \label{fig:permutation-test}
\end{figure}

We compare the absolute temperature distributions of the peak 5-day heatwaves occurring in each 10-year period (note that these events are not temporally aligned across the different datasets). Although the RCM ensemble mean exhibits a slight negative temperature bias during peak heatwave days relative to the emulator (see \Cref{fig:combined-paris-plots}a), the histogram shows that all RCM heatwaves are fully spanned by the EnScale ensemble spread (\Cref{fig:combined-paris-plots}b). Further, a two-sided permutation test ($N=10^4$) suggests that this discrepancy is not statistically significant ($p=0.94$; \Cref{fig:permutation-test}.

\begin{figure}[H]
    \centering
    \includegraphics[width=\linewidth]{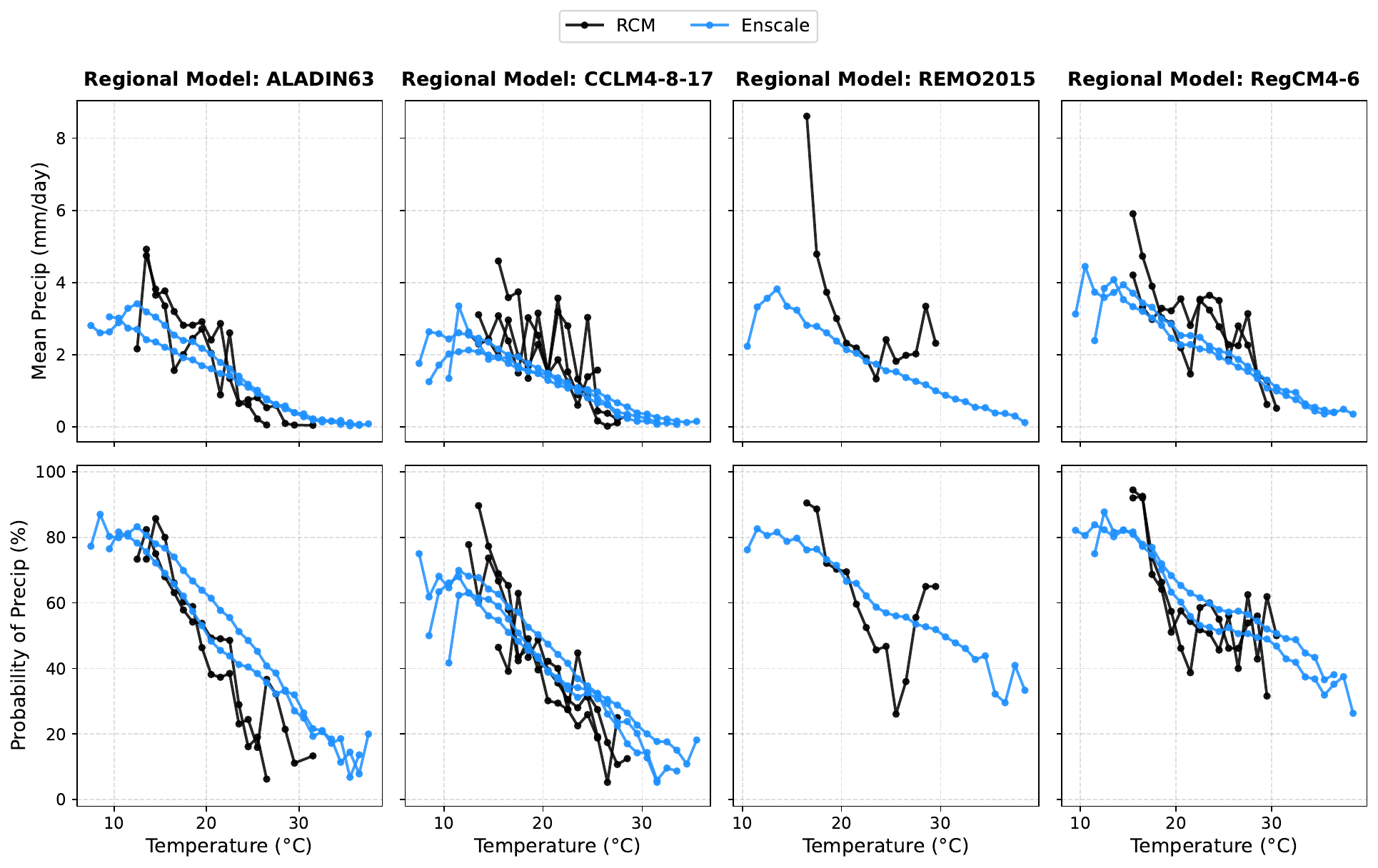}
    \caption{Temperature dependence of daily precipitation characteristics at the Paris grid cell during summer across model classes, subset by RCM configuration. Top subpanels depict mean daily precipitation amount (mm/day) binned in $1.0\,^\circ\text{C}$ temperature intervals; bottom subpanels show precipitation probability (percentage of days with $>0.1\,\text{mm/day}$).}
    \label{fig:sep-by-rcm}
\end{figure}

\begin{table}[H]
\centering
\caption{Weighted log-linear probability decay rates ($\beta$) of $1\sigma$ heatwave event durations at the Paris grid cell across model classes. Heatwave duration is defined as the number of consecutive days where daily mean temperature exceeds $1\sigma$ above the reference summer climatology. Slopes ($\beta$, in $\text{day}^{-1}$) are estimated via weighted least-squares (WLS) regression, using raw event counts as weights. Two-sided $p$-values assess slope difference interaction terms relative to the reference \texttt{EnScale} model.}
\label{tab:prob_slope_results}
\begin{tabular}{lS[table-format=-1.4]cc}
\hline
 & \textbf{Decay Slope ($\beta$)}  & \textbf{$p$-value} \\ \hline
EnScale & -0.2773 &  \\
RCM      & -0.3356 & 0.601 \\
GCM       & -0.3040 & 0.898 \\ \hline
\end{tabular}
\end{table}

\begin{figure}[H]
    \centering
    \includegraphics[width=\linewidth]{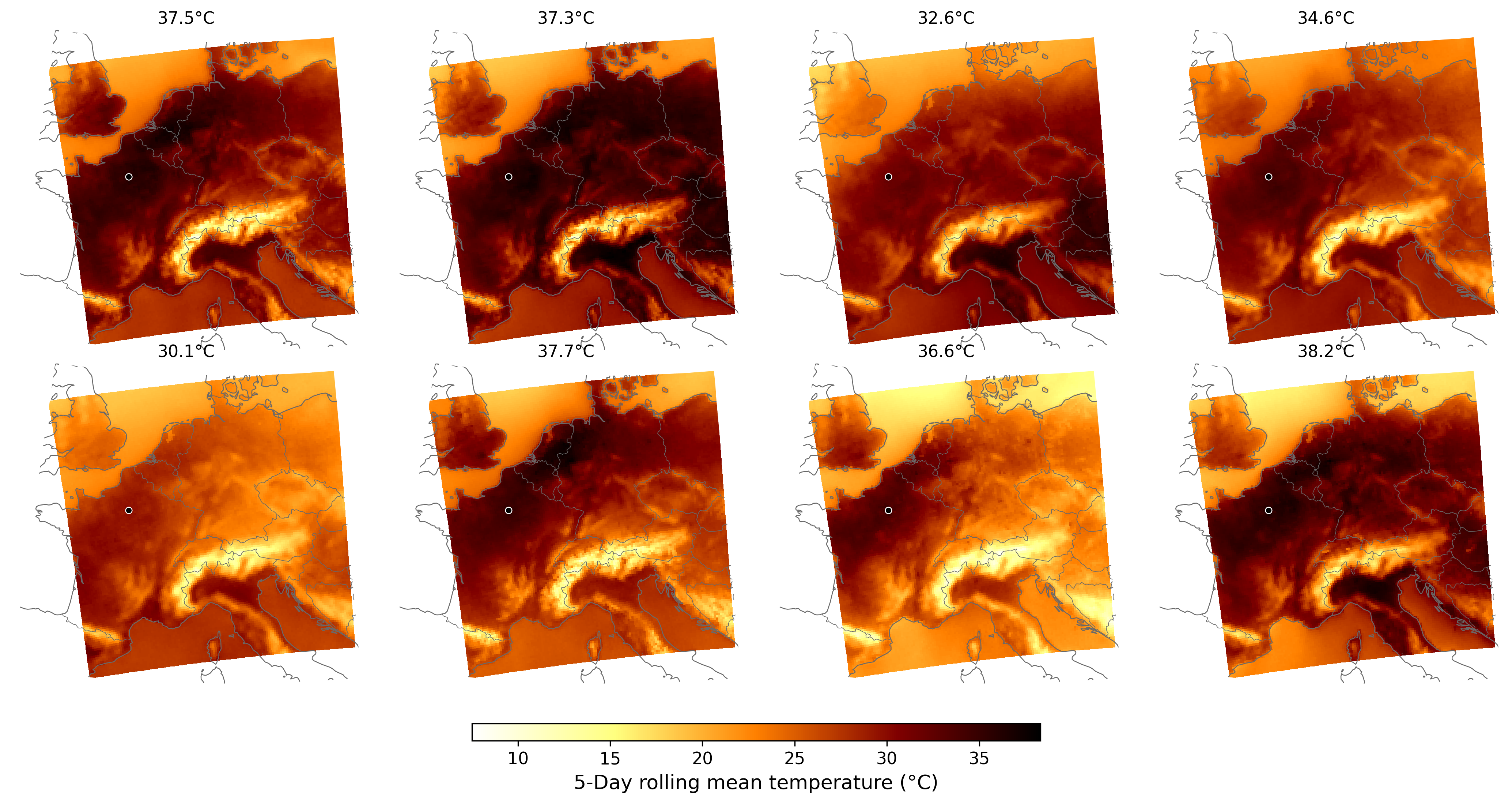}
    \caption{Spatial patterns of the peak 5-day rolling mean temperature ($^\circ\text{C}$) during the absolute hottest summer event in Paris (black 'x') per EnScale model. Subpanel titles indicate the peak 5-day average temperature in Paris. The general patterns look similar with the heatwave covering a larger area.}
    \label{fig:spatial-max}
\end{figure}

\begin{figure}[H]
    \centering
        \includegraphics[width=\linewidth]{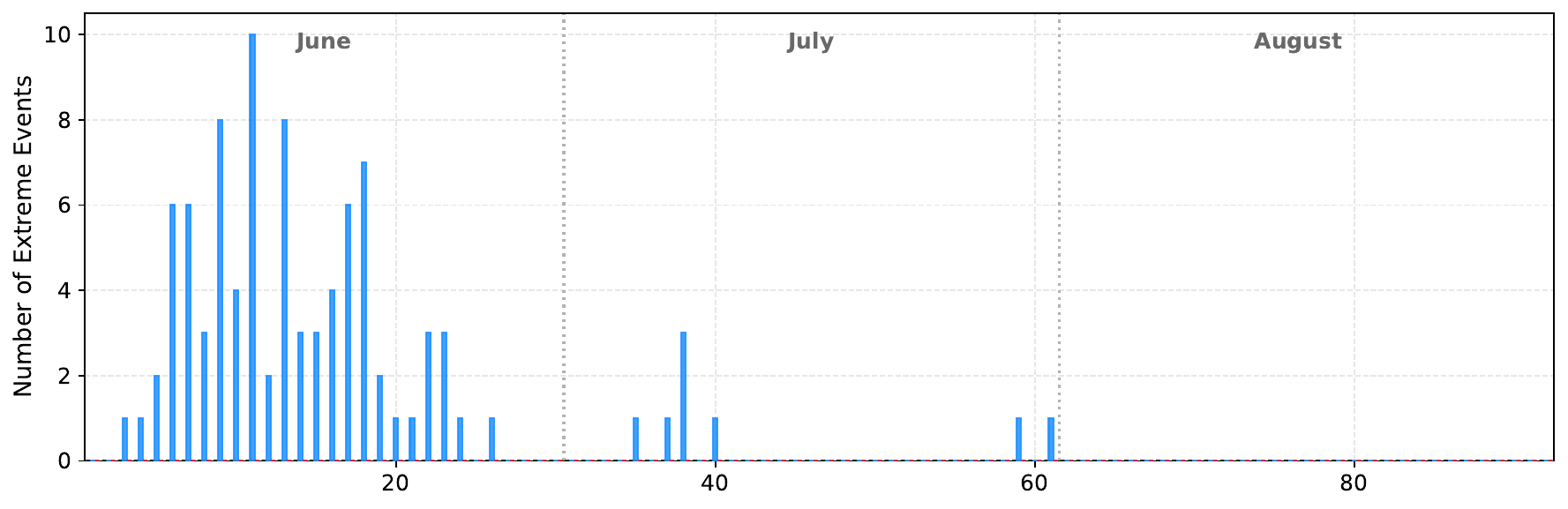}
        \label{fig:timing-high-sigma}
    \caption{Temporal distribution of extreme 5-day heatwave events exceeding a $4.6\sigma$ standardized anomaly threshold at the Paris grid cell during summer, indexed by day of summer. Standardized anomalies ($\sigma$) are calculated using 5-day centered rolling mean temperatures relative to the reference pooled summer climatology mean and variance. }
    \label{fig:temporal-distribution}
\end{figure}

\begin{figure}
    \centering
    \includegraphics[width=\linewidth]{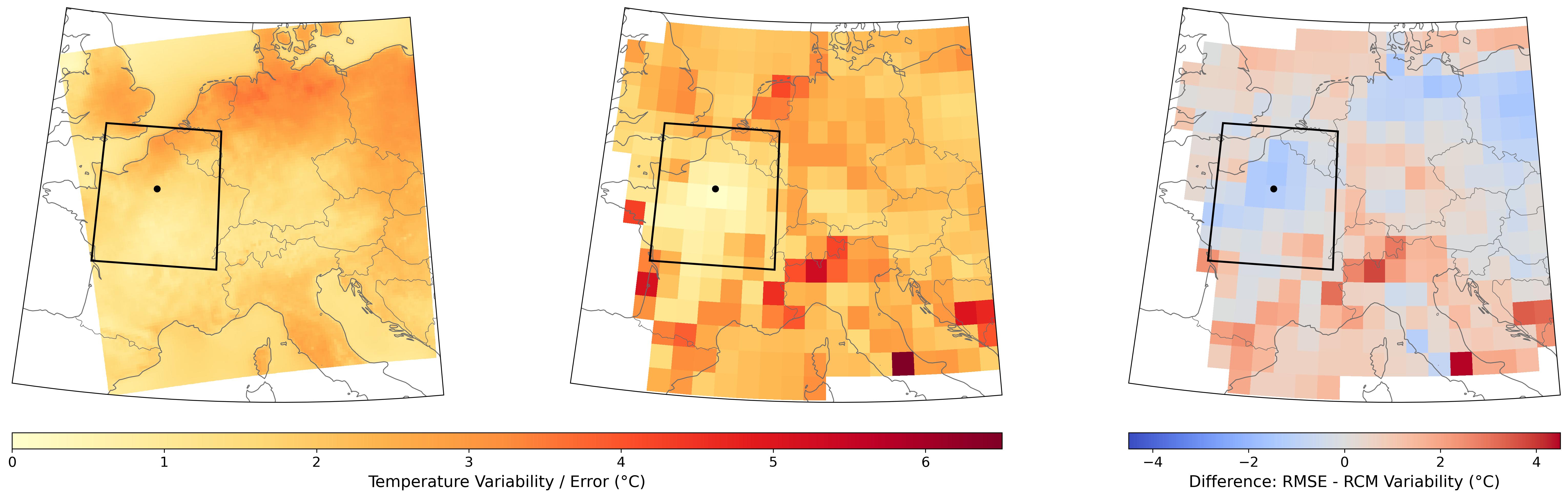}
    \caption{Spatial comparison of RCM ensemble variability against EnScale analogue-matching error across Europe. Left: RCM model variability, computed as the pixel-wise standard deviation across the regional climate model ensemble for the most extreme heatwave event per model. Center: EnScale spatial downscaling mismatch, calculated as the pixel-wise root-mean-square error between the coarse target fields and the analogs selected via a two-step matching procedure. Right: Spatial difference field ($\mathrm{RMSE} - \mathrm{RCM\ Variability}$). Blue regions highlight areas where the downscaling error is lower than the multi-model RCM spread. The black marker denotes the location of Paris, and the black rectangle outlines the regional $7 \times 7$ grid-cell evaluation window ($\pm 3$ GCM cells around Paris).}
    \label{fig:var_vs_rmse}
\end{figure}

\end{appendices}
\end{document}